# Nonlinear vibration of a dipteran flight robot system with rotational geometric nonlinearity

Yanwei Han[a], Zijian Zhang[b,*]

[a] School of Civil Engineering and Architecture, Henan University and Science and Technology, Luoyang 471023, China;

[b] College of Astronautics, Nanjing University of Aeronautics and Astronautics, Nanjing 210016, China

* Corresponding author. E-mail address: zj.zhang@nuaa.edu.cn (Z. Zhang)

**Abstract:** The dipteran flight mechanism of the insects is commonly used to design the nonlinear flight robot system. However, the dynamic response of the click mechanism of the nonlinear robot system with multiple stability still unclear. In this paper, a novel dipteran robot model with click mechanism proposed based on the multiple stability of snap-through buckling. The motion of equation of the nonlinear flight robot system is obtained by using the Euler-Lagrange equation. The nonlinear potential energy, the elastic force, equilibrium bifurcation, as well as equilibrium stability are investigated to show the multiple stability characteristics. The transient sets of bifurcation and persistent set of regions in the system parameter plane and the corresponding phase portraits are obtained with multiple stability of single and double well behaviors. Then, the periodic free vibration response are defined by the analytical solution of three kinds of elliptical functions, as well as the amplitude frequency responses are investigated by numerical integration. Based on the topological equivalent method, the chaotic thresholds of the homo-clinic orbits for the chaotic vibration of harmonic forced robot system are derived to show the chaotic parametric condition. Finally, the prototype of nonlinear flapping robot is manufactured and the experimental system is setup. The nonlinear static moment of force curves, periodic response and dynamic flight vibration of dipteran robot system are carried out. It is shown that the test results are agree well with the theoretical analysis and numerical simulation. Those result have the potential application for the structure design of the efficient flight robot.

**Key Words**: Dipteran robot system; equilibrium bifurcation and stability; periodic solutions; amplitude frequency; chaotic threshold.

## 1. Introduction

The dipteran insects of butterflies, dragonflies and mosquitoes with the double flapping wings are widely distributed worldwide, such as in the tropics, arctic, oceans, lakes, and high mountains [1]. There are some debate on the nonlinear mechanism of how excellent flight performance of flight motor of diptera. As shown in Fig. 1(a), Leonardo da Vinci designed a variety of aircraft and gliders to realize his dream of flying [2]. During the last decade, the flight mechanism of diptera have been proposed and studied with nonlinear restoring and linear or quadratic damping force. As illustrated in Fig. 1(b), many researchers are trying to reveal the flight mechanism of the dipteran insect by methods of theoretical biology and view of dynamical mechanics [3]. The bionic flapping wing aircraft is a flight mode that imitates the flapping wings of birds and other creatures. Such a flight mode has the potential to achieve long-distance flight at the scale of micro and small aircraft. At the same time, flapping wings have more freedom of movement and strong maneuverability. Not only that, bionic flapping wing aircraft has the characteristics of bionics, concealment and portability, which are generally hand throwing takeoff, gliding and landing. The flapping robot not only could take off and landing without limited by the site, but also can achieve rapid high flight, long-distance cruise, etc. like micro and small fixed wing aircraft. Recently, the flapping insect flight performance and click mechanism have been widely investigated by methods of theoretical analysis, numerical simulation and experimental test.

Firstly, the rigid body dynamics of the configuration design of flight robot are extensively investigated. Thomson et al. analyzed the kinematic characteristics of the wing beat in dipteran flight model by using of the anatomical and physiological parameters, mathematical model and the catastrophe theory [4]. Branan et al. presented the dynamic analysis of a simplified model with the flight "click" mechanism which exhibited distinct advantages over a system at the resonant frequency [5]. Lau et al. developed the compliant design of thoracic mechanism by mimicking a dipteran insect flight thorax with nonlinear stiffness characteristics which can saves the inertial power of the micro-air vehicles of the flapping wing[6]. Gunther et al. presented a novel method of the parallel elastic mechanism which is able to carry payloads at least three times of its body weight and realize stable forward hopping [7]. Abas et al. reviewed the application of the piezoelectric transmission to replace the conventional motorized transmission for the nano-aerial vehicles (NAV) and the micro-aerial vehicles (MAV) flapping wing motion [8]. Jankauski found the nonlinear characteristic of the thorax of the flapping wing micro-air vehicles with the hardening spring to increase flapping frequency bandwidth and to build a foundation for future research [9]. Lietz et al. firstly presented the systematic study of the wing damping of the flight flying insects and shown that different wing regions have almost identical damping properties which is essential to maintain stability of the orientation and direction of motion [10]. Combs et al. developed a simple mathematical method to estimate the spatial flexural stiffness variation of the insect wings [11]. Duncan et al. investigated a viable alternative of an parallel-elastic actuation strategy for the agile robots by using series-elastic power modulation [12]. Hunt et al. proposed a parallel elastic hopping robot with the control input of the pulsed signals outperforms the sinusoidal input with a lower energy expenditure [13]. Although the mechanical dynamic features of the flight mechanism are studied many researchers in modelling, computation and testing, but the accurate dynamic model has never been established yet. Therefore, this paper may provide an archetypal precise model for the dipteran flight robot system and the proposed mathematical model can accurately describe the vibration response of wings.

On the other hand, the flight fluid dynamics of the dipteran flight system have been researched extensively. Somps et al. successfully revealed that the novel mechanisms of insect flight of the large aerodynamic lift force comes from the unsteady flow-wing interaction [14]. Warrick et al. thought the aerodynamic mechanisms of hummingbirds which is similar to the insects despite profound musculoskeletal differences [15]. Bomphrey et al. reported the trailing-edge vortices, leading-edge vorticesand rotational drag of the free-flight mosquito wing kinematics by using the computational fluid dynamics [16]. Muijres et al. studied the escape speed and the wing forces in the malaria mosquito using the aerodynamic modeling during the push-off phase [17]. Nakata et al. clarified the aerodynamic effect of the flapping wing deformation by means of the computational fluid dynamics [18]. Dickinson et al. found the enhanced aerodynamic performance of the flight insects comes from the interactive mechanisms of the wake capture, the rotational circulation as well as the delayed stall [19]. Johansson et al. suggested the high flight power of the hawmoths that was influced by force production of the strong, complex and variable leading edge vortex (LEV) structure [20]. Lehmann et al. presented an unusual insect flying mechanism that can effectively harvest the rotational motion energy from the surrounding air vortices which can reduce fight power of energetic expenditures [21-22]. Chin et al. developed a low-loss anti-whirl transmission of flapping wings of the performance of aggressive flight by simple tail control for the roles of the propulsion, lift and drag [23]. While detailed researches had been conducted on aerodynamics using theoretical and experimental methods, there is still relatively little systematic research on fluid structure coupling. Therefore, we develop an archetypal mathematical modelling to accurately describe the dipteran fly mechanism.

Lastly, the bio-mechanical investigations of the flight insects and the jumping animal are generally discussed. Heitler et al. investigated the locust jumps with a rapid extension of the metathoraeic tibiae by the tension isometrically and the co-contracting with the flexor muscle [24]. Iwamoto et al. found that the insects use a refined preexisting force-enhancing mechanism of the flight muscle-specific features or preexisting contractile functions to realize the high-frequency wing beat [25]. Ilton et al. presented the synthetic and biological model of the spring and latch dynamics to reveal a foundational framework of power-amplified systems for the scaling, synthetic design as well as evolutionary diversification [26]. Steinhardt et al. provided the temporal phase characteristics of the robotic, mathematical and biological systems to understand the function of linkages dynamics and latches structure [27]. Harne et al. used a representative structure of the biological model to reveal the flight mechanism of the axial support stiffness and compression characteristics and to modulate the amplitude range and the wing stroke dynamics [28]. Majumdar et al. numerically investigated the fluid forces and dynamical states of the passive pitching-plunging flapping foil to avoid the aperiodic transition under the high plunge velocity [29]. Although many researchers analyze flight dynamics from a biomechanical perspective, their mechanical and mathematical mechanisms have not been proposed. Hence, the proposal of a key and precise dynamic model has become urgent to solve.

The rest of paper is outlined as follows. In Section 2, the equation of motion is derived by Lagrange method. In Section 3, the nonlinear force, potential energy and phase portraits are plotted to shown the varying stiffness and multiple stability. In Section 4, the equilibrium bifurcation and stability are investigated to show the transition behavior. In Section 5, the amplitude frequency response of the free, forced and chaotic oscillation are studied to exhibit the complex dynamic response. In Section 6, experimental work is carried out to verified the theoretical and simulation result. Lastly, Section 7 conclude with the main finding and outlook of the future work.

## 2. Modeling of flapping wing system

### 2.1 Flapping wing model

As illustrated in Fig. 1(a), the drawings of the world's first flapping wing aircraft was designed by a literary and scientific giant of Leonardo da Vinci at the beginning of the 15th century of the European Renaissance. He studied bird wings and used physical and anatomical knowledge and his drawings and drafts are still well preserved in the museum [1].

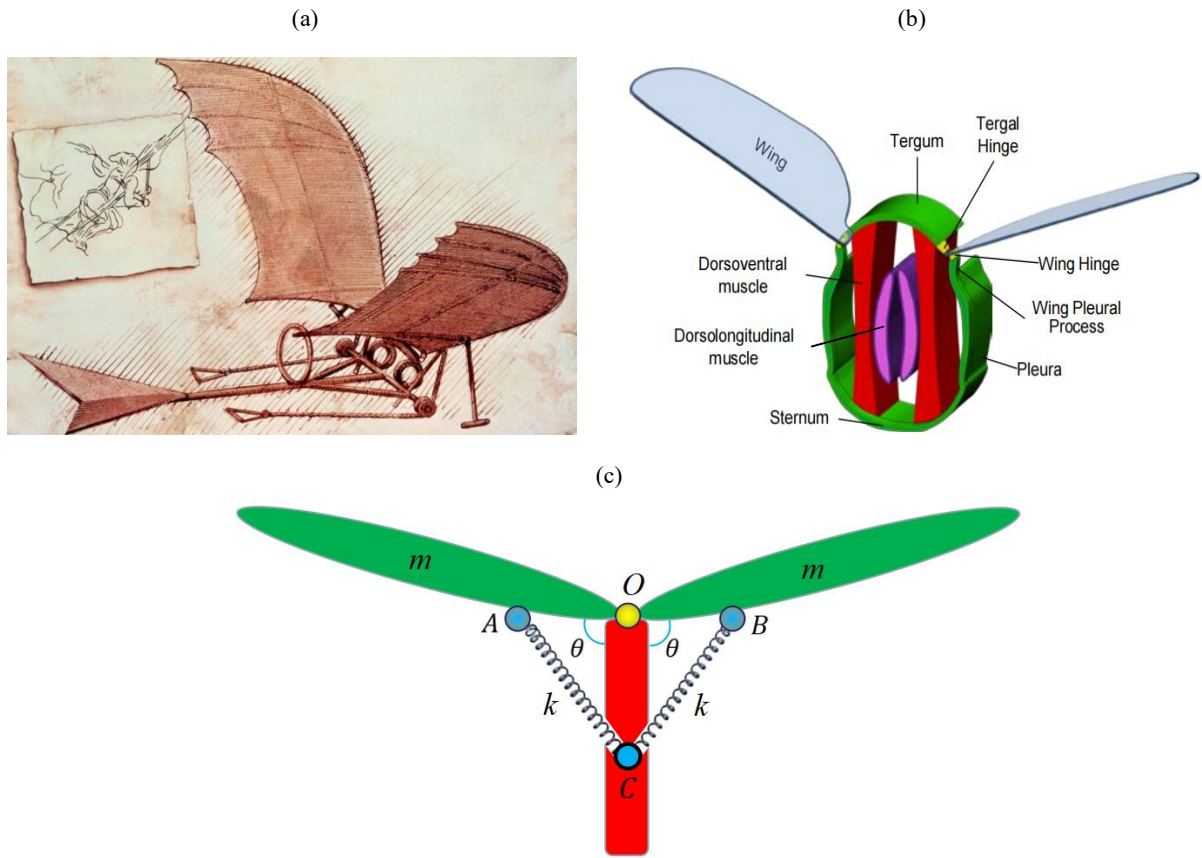

Fig. 1 Modeling of flapping robot system. (a) Flapping wing aircraft model designed by Leonardo da Vinci [1]. (b) Physical model of the flapping flight insect system driven by asynchronous indirect muscles [2-3]. (c) Mechanical model of flapping robot [4].
(Color online)

As indicated in Fig. 1(b), work way of two antagonistic flight muscles in the thorax, dorsal longitudinal muscle (DLM) and dorsoventral muscle (DVM) is that when one contracts while the other is stretched. Consequently, those two muscles alternately stretch each other to cause sinusoidal actuator and to enable continuous wing-beats. The upstroke of wings are driven by the dorsal vertical flight muscles pull on the thorax roof while stretching the dorsal longitudinal muscles. Whereas the downstroke of wings results from the contraction of dorsal longitudinal muscles of the shorten posterior ends of the thorax and the stretching of the dorsal vertical flight muscles. Therefore, wing movement of fruit fly driven by asynchronous indirect flight muscles periodically upstroke and downstroke by the cycle stretch and contraction of muscles [3].

As shown in Fig. 1(c), the flapping robot model consist of mass, spring, damper and external force is presented based on the flight motor [35]. The nonlinear elastic force come from the geometrical nonlinearity of spring structure. The elastic force exhibits the flapping wing mechanism. $m$ (kg) is the seismic mass of wings which attached to the joint points $A$ and $B$. $\theta$ (rad) is the generalized angle coordinate of lines the lines $OA$ and $OC$ in direction of clockwise rotation, as well as the lines $OB$ and $OC$ in direction of counter clockwise rotation around the point $O$. Here, we suppose the counter clockwise of the angular displacement is the positive direction. $k$ (N/m) and $l$ (m) denote the linear stiffness constant and the free length of elastic linear springs respectively, which can be elongated or reduced and linked two points $A$ and $C$ or $B$ and $C$. $c$ (N·s/m) is the linear viscous damping, which hinged at two points $A$ and $C$ or $B$ and $C$. $a$ (m) and $b$ (m) are the length of the joint points between the hinges point $OA$ and $OB$.

Table 1. Parameters of flapping robot system

| Parameter | Symbol | SI Unit | Dimension |
| --- | --- | --- | --- |
| Lumped mass of wings | $m$ | Kg | M |
| Stiffness of springs | $k$ | N/m | $MT^{-2}$ |
| Damping coefficient | $c$ | N·m/rad | $ML^2T^{-2}$ |
| Angular of rotation | $\theta$ | Rad | 1 |
| Free length of oblique springs | $l$ | m | L |
| Length of $OA$ and $OB$ | $a$ | m | L |
| Length of $OC$ | $b$ | m | L |
| The radius of inertia of wings | $d$ | m | L |
| Amplitude of excitation moment | $m_0$ | N·m | $ML^2T^{-2}$ |
| Frequency of the external excitation | $\omega_0$ | rad/s | $T^{-1}$ |
| Dimensional time | $t$ | s | T |

## 2.2 Equation of motion

Conventionally, using the included angle $\theta$ as the generalized coordinate for the robot model shown in Fig. 1 (c), the total rotational kinetic energy $KE$ (J) stored in the mass by virtue of the rotational velocity for the nonlinear flapping robot system is given by the follows

$$KE = I\dot{\theta}^2 \tag{1}$$

where $I = 2md^2$ is the polar mass moment of inertia about the hinge point $O$ and $\dot{\theta} = d\theta/dt$ (rad/s) denotes the angular velocity that rotated at the hinge point $O$ and the overdot represents the differentiation with dimensional time $t$.

Since the potential energy $PE$ (J) is stored in the springs by virtue of its elastic deformation of the strain referred to its lowest energy position, and then the nonlinear total potential energy for the flapping robot system can be written in following form [30]

$$PE = k\left(\sqrt{a+b-2ab\cos\theta} - l\right)^2 + 2mgd \tag{2}$$

here $a$, $b$ and $l$ are the length of $OA$, distance of $OB$ and the uninstructed springs respectively. $g$ (m/s²) is the acceleration of the gravitational acceleration. Although the spring themselves are the elastic linear, the resulting force that acted on the mass is a strong nonlinear by reason of the geometrical configuration.

Moreover, the damping is unavoidable in real system, which leading to the dissipation of mechanical energy. So the nonlinear dissipation function $\Psi$ (J/s) of the mechanical vibration of the flapping wing robot are obtained as following

$$\Psi = \frac{1}{2}\frac{c\dot{\theta}^2(ab\sin\theta)^2}{a^2+b^2-2ab\cos\theta} \tag{3}$$

herein $\Psi$ is the mechanical damping and the overdot represents differentiation with respect to time $t$. The Eq. (3) are function of the angular $\theta$ and the angular velocity $\dot{\theta}$, called Rayleigh damping.

Furthermore, the external forces are widely presented in the practical vibration engineer. Hence the generalized external force $Q$ (N·m) of the gravitation of gravitation $g$ (m/s²) and sinusoidal exciting forces is given by

$$Q = 2m_0 \sin(\omega t + \varphi) \quad (4)$$

in which $m_0$, $\omega$ and $\varphi$ are the external moment of force amplitude, the driving frequency and the phase respectively.

Subsequently, the govern equitation can be obtained by using the method of the Euler-Lagrange equations in the following form

$$\frac{d}{dt}\left(\frac{\partial \Pi}{\partial \dot{\theta}}\right) - \frac{\partial \Pi}{\partial \theta} + \frac{\partial \Psi}{\partial \theta} = Q \quad (5)$$

where $\Pi = KE - PE$ is Lagrange function.

By continuity, with the expressions of the kinetic energy Eq. (1), The potential energy Eqs. (2), damping function of Eq. (3), as well as the external moment of force Eq. (4), we obtain the differential governing equation of the flapping wing robot system with geometrical non-linearity, based on the Lagrange Eq. (5), as follow form

$$I\ddot{\theta} + \frac{c\dot{\theta}(ab\sin\theta)^2}{a^2 + b^2 - 2ab\cos\theta} + \left(kab\left(1 - \frac{l}{\sqrt{a^2 + b^2 - 2ab\cos\theta}}\right) + 2mg\right)\sin\theta = m_0 \sin(\omega_0 t + \varphi) \quad (6)$$

where $I = 2md^2$ is the moment of inertia and Eq. (6) can be also obtained by applying Newton's law for mechanical system.

For convenient investigation of the nonlinear robot system (6), the dimensionless transformation of the geometrical and physical parameters are defined as following form

$$\alpha = \frac{a}{l},\ \beta = \frac{b}{l},\ \gamma = \frac{2mg}{kl^2},\ \kappa = \frac{I}{ml^2},\ \omega_n = \sqrt{\frac{k}{m}},\ \xi = \frac{c}{2\sqrt{mk}},\ M_0 = \frac{m_0}{kl^2},\ \Omega_0 = \frac{\omega_0}{\omega_n} \quad (7)$$

where $\alpha$ and $\beta$ are the non-dimensional geometrical factors, $\gamma$ is the dimensionless coefficient of gravitation, $\kappa$ is the ratio of moment of inertia, $\omega_n$ is the natural frequency depending on the mass and stiffness, $\xi$ is the ratio of damping associated with mass, stiffness and damper. $M_0$ is the dimensionless torque of the external excitation force, $\Omega_0$ is the ratio of the forced non-dimensional frequency.

Similarly, aim at the convenient study of the dynamical robot system (6), we define dimensionless version of the angle $\theta$ and the time $t$ as follows

$$\theta = \frac{\theta kl}{mg},\ T = \omega_n t \quad (8)$$

where $\theta$ denotes the non-dimensional angle and $t$ represents the new independent variable dimensionless time.

With help of the transformation expression of Eqs. (7) and (8), the nonlinear flapping robot system (6) are transformed into the non-dimensional form as follows

$$\kappa\ddot{\theta} + \frac{2\xi(\alpha\beta\sin\theta)^2}{\alpha^2 + \beta^2 - 2\alpha\beta\cos\theta}\dot{\theta} + \left(\gamma + \alpha\beta\left(1 - \frac{1}{\sqrt{\alpha^2 + \beta^2 - 2\alpha\beta\cos\theta}}\right)\right)\sin\theta = M_0 \sin(\Omega_0 T + \varphi) \quad (9)$$

For convenience, by introducing the new dependent variable $\dot{\theta} = \omega$, a two dimensional first-order equations of perturbed dipteran flight robot system is convenient to replace Eq. (10) by

$$\begin{cases} \dot{\theta} = \omega \\ \kappa\dot{\omega} = -\frac{2\xi\omega(\alpha\beta\sin\theta)^2}{\alpha^2 + \beta^2 - 2\alpha\beta\cos\theta} - \left(\gamma + \alpha\beta\left(1 - \frac{1}{\sqrt{\alpha^2 + \beta^2 - 2\alpha\beta\cos\theta}}\right)\right)\sin\theta + M_0 \sin(\Omega_0 T + \varphi) \end{cases} \quad (10)$$

where $\omega$ is the dimensionless angular velocity.

In particular, for the parameter setting of $\xi = 0$ and $M_0 = 0$, a pair of first-order ordinary equations of the unperturbed dipteran flight robot system in replace of Eq. (9), we can write

$$\begin{cases} \dot{\theta} = \omega \\ \kappa\dot{\omega} = -\left(\alpha\beta\left(1 - \dfrac{1}{\sqrt{\alpha^2 + \beta^2 - 2\alpha\beta\cos\theta}}\right) + \gamma\right)\sin\theta \end{cases} \quad (11)$$

where $\omega$ is the dimensionless angular velocity. In generally, system (11) cannot be analytically solved due to the difficult nonlinearity of radical, square, root square and harmonic function. However, Qualitative results of the autonomous system (11) shown that the proposed system exhibited complex response, such as, pitchfork bifurcation, the singular points of the centers $(\theta_1, 0)$, $(\theta_3, 0)$, $(\theta_4, 0)$ and saddle $(\theta_2, 0)$, as well as the closed trajectories of the periodic solution and the aperiodic hetero-clinic orbit.

## 3. The static property analysis

In this section, the static nonlinear response of the potential energy, Hamilton function, equilibrium bifurcation are studied by the qualitative methods of the nonlinear vibration theory.

### 3.1 Potential energy

By integral calculus of the Eq. (11), the nonlinear potential energy function of the free vibration of the dipteran flight robot may be expressed as

$$PEN = 0.5\left(\sqrt{\alpha^2 + \beta^2 - 2\alpha\beta\cos\theta} - 1\right)^2 + \gamma(1 - \cos\theta) \quad (12)$$

herein $\theta$ is the general coordinate and real number, $\alpha$ and $\beta$ are the positive real number. It is found that potential energy function $PEN$ include both irrational term and harmonic term, which can not be solved by the conventional approach. Therefore, the those barriers may cause the analytical difficulty of both static and dynamic response, which will be studied in the subsequent text.

As shown in Fig. 2, the nonlinear potential energy diagrams of flapping wing robot system are plotted in the three dimensional $\theta - \alpha - PEN$ parametric space and two dimensional $\theta - PEN$ parametric plane. For fixed gravitation parameter $\gamma = 0.5$, the surface and curves for different value of geometrical parameter $\alpha$ are obtained in Figs. 2(a, b). As shown in Figs. 2 (c, d), the surface and curves for different value of geometrical parameter $\alpha$ are obtained for parameter setting of $\gamma = 0.5$. Saddle points of unstable equilibrium positions corresponding to the maxima of the the potential energy $PEN$, as well as the centers corresponding to the minima of the potential energy $PEN$.

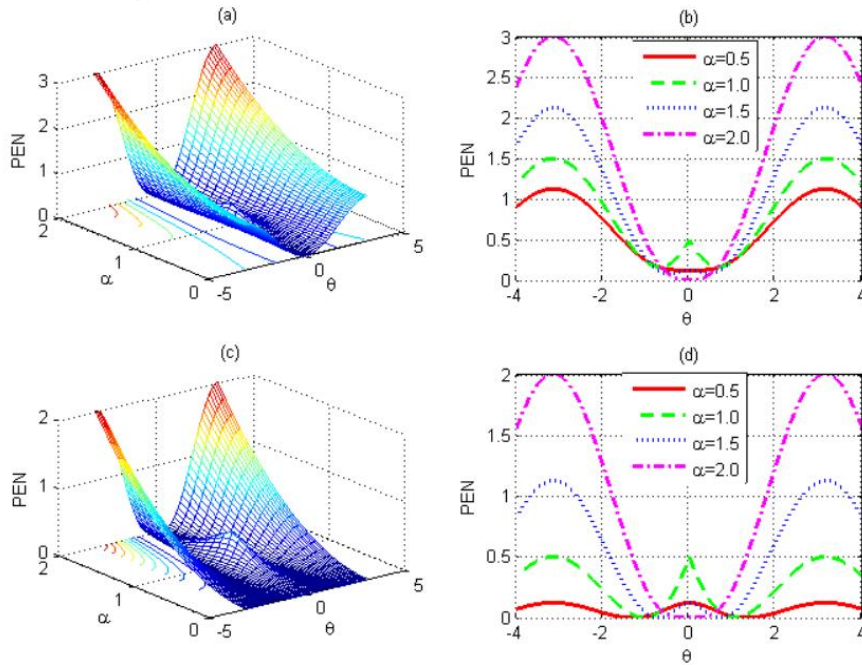

Fig. 2 The nonlinear potential energy for $\beta = 1$. (a) The curved energy surface in the tridimensional parameter $\theta - \alpha - PEN$ space and (b) the nonlinear energy curves on bidimensional $\theta - PEN$ plane for different value of geometrical parameter $\alpha$ and with $\gamma = 0.5$. (c) Energy surface and (d) curves for different value of geometrical parameter $\alpha$ and with $\gamma = 0$

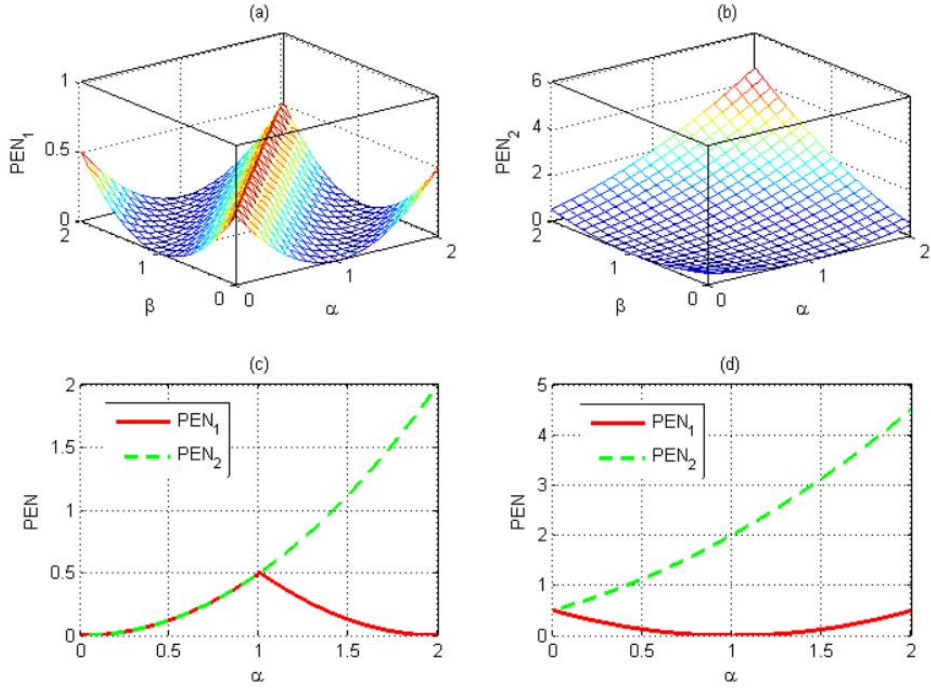

Fig. 3 The maximum potential energy $PEN$ curved surface for the gravitational ratio $\gamma = 0$. (a) The dimnsionless potential energy function $PEN_1$ curved surfaces in three dimensional $\alpha - \beta - PEN$ parameter space. (b) The potential energy $PEN_2$ of Eq. (13). (c) The nonlinear curves of potential energy $PEN$ for the geometrical coefficient $\beta = 1$ and (d) $\beta = 2$.

In particular, for the gravitational ratio setting of $\gamma = 0$, the nonlinear potential energy at equilibrium angle for the unperturbed robot system (11) are defined as following

$$\begin{cases} PEN_1 = 0.5(1-|\alpha-\beta|)^2 \\ PEN_2 = 0.5(1-|\alpha+\beta|)^2 \end{cases} \quad (13)$$

where $PEN_1 = PEN(\theta = 0)$ and $PEN_2 = PEN(\theta = \pm \pi)$.

As shown in Figs. 3(a, b), the maximum $PEN_1$ and $PEN_2$ surfaces are given to exhibit the nonlinear relationships between the potential energy $PEN$ and system parameters $\alpha$ and $\beta$. In Fig. 3(c), the maximum of $PEN_1$ and $PEN_2$ are the same when $\alpha$ in the region of (0, 1) and $PEN_2$ is bigger than $PEN_1$ for $\alpha$ range from 1 to 2. As presented in Fig. 3(d), the values of $PEN_2$ is always bigger than $PEN_1$ for the geometrical coefficient $\alpha$ range from 0 to 2.

**3.2 Nonlinear moment of force**

The nonlinear moment of the elastic springs force rotating about the points O for the free vibration of the flight robot system (11) is denoted by [31]

$$M_s = \alpha\beta\sin\theta\left(1-1/\sqrt{\alpha^2+\beta^2-2\alpha\beta\cos\theta}\right)+\gamma\sin\theta \quad (13)$$

where $M_s$ is the dimensionless formula of moment of force.

As shown in Figs. 4 (a, c), the non-dimensional moment of force $M_s$ as function of both the angle $\theta$ and geometrical parameter $\alpha$ are plotted in a three dimensional $\theta - \alpha - M_s$ parameter space. It is found that the gradient of curved moment surface $M_s$ is nonlinearly dependent upon the value of the angle variable $\theta$, as well as geometrical coefficient $\alpha$. For $\beta = 1$, when the $\alpha$ less than the value 2, then the nonlinear moment (13) exhibits both positive and negative stiffness depending upon the non-dimensional angle $\theta$ which lead to the click mechanism or the snap through behavior [32].

As illustrated in Fig. 4(b, d) of two-dimensional graph, the stiffness is always positive and there is no click mechanism when $\alpha$ is greater than 2. It is can be seen that the stiffness increase with increasing positive and negative angle about $\theta = 0$ and the flight robot system (11) exhibits the hardening stiffness spring for the geometrical parameter setting $\alpha \geq 2$. Therefore, it is concluded that the click mechanism can be switch on or off by adjusting the geometrical parameters $\alpha$ and $\beta$. In this paper, the interesting and important case of the configuration of geometrical parameter $0 < \alpha < 2$ are investigated in detail.

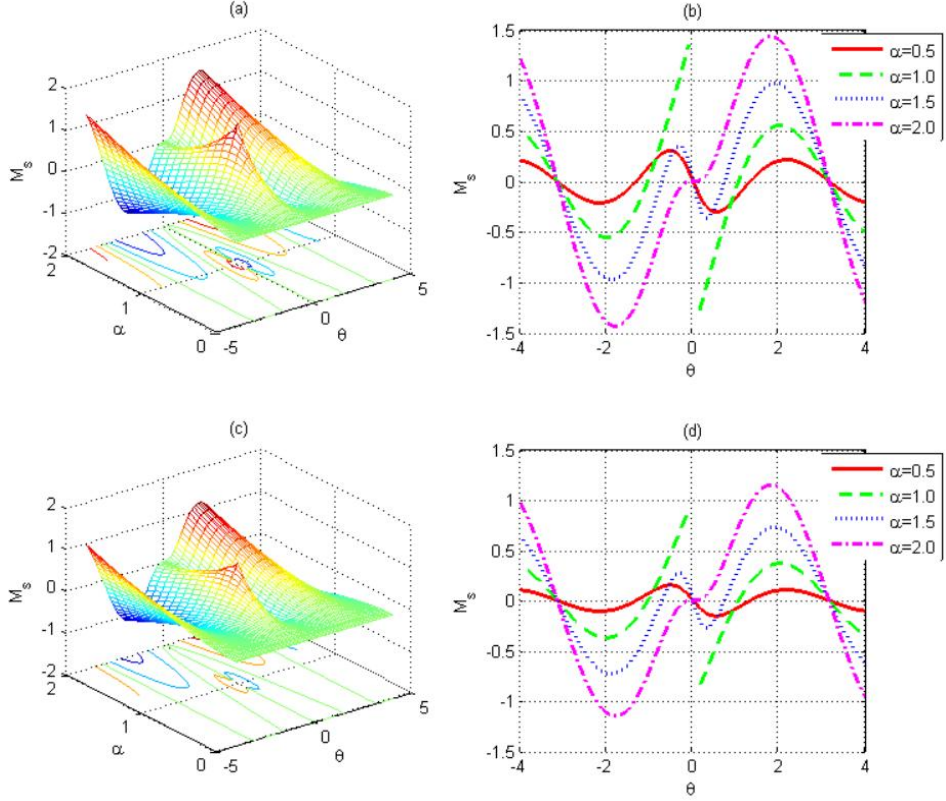

Fig. 4 Nonlinear static torsional moment of force as function of the angle $\theta$ and geometrical parameter $\alpha$ for Eq. (13). (a) The curved moment surface in the tridimensional $\theta - \alpha - M_s$ parameter space with $\beta = 0.5$. (b) The moment of force curves on the two dimensional parameter plane $\theta - M_s$ with $\beta = 0.5$. (c)The curved moment surface with $\beta = 0$. (d) The moment of force curves with $\beta = 0$.

### 3.3 Hamilton function and phase portraits

Hamilton function of the total mechanical energy corresponds to the sum of kinetic energy and potential energy for the free vibration of flight robot system (11) is given by

$$H = \frac{1}{2}\kappa\omega^2 + \frac{1}{2}\left(\sqrt{\alpha^2 + \beta^2 - 2\alpha\beta\cos\theta} - 1\right)^2 + \gamma(1 - \cos\theta) \tag{14}$$

where $\kappa, \alpha, \beta$ and $\gamma \geq 0$ are the positive real number, $\theta$ represents the included angle and $\omega$ denotes the dimensionless angular velocity. It is noted that the system can display the transition behavior from single well to double well dynamics depending on the geometrical coefficient $\alpha, \beta$ and the gravitational ratio $\gamma$.

For the unperturbed robot system (11), the small periodic solution center at fixed point $(\theta_{1,2}, 0)$ of two stable static equilibrium positions corresponding to the Hamilton value of $0 < H < PEN_1$ for $0 < \alpha \leq 1$ and $\beta = 1$. The bigger periodic solutions corresponding to the Hamilton value of $H > PEN_1$ for $0 < \alpha \leq 1$ and $\beta = 1$. The bifurcation and catastrophe theory provides a mathematical tool for described multiple system with the click mechanism and snap through phenomena.

As illustrated in Fig. 5, the structure of phase portraits near the centers $(\theta_2, 0)$ are radically altered under the external perturbation. From Fig. 5, it is shown that the non-hyperbolic fixed points $(\theta_2, 0)$ with positive stiffness of the centers surrounded by a family of periodic orbits, with the increasing frequency $\Omega_n$ of increasing Hamilton from 0 to $H_0 = (|\alpha \mp \beta| \pm 1)^2 + \gamma$. It is also found that the hyperbolic fixed points $(\theta_1, 0)$ and $(\theta_2, 0)$ is the saddle with negative stiffness.

As shown in Figs. 5 (a, b), it is found that the system (11) exhibits the double well and heteroclinic orbits connected the saddle point $(\theta_1, 0)$ and $(\theta_2, 0)$. Moreover, there are the periodic trajectories around the equilibrium centers $(\theta_3, 0)$ and $(\theta_4, 0)$ for Hamilton value of $0 < H < 0.125$. It can be seen that the dashed line denotes the small periodic solution, the dash-dotted lines represent the heteroclinic orbits, as well as the solid lines indicate the big trajectories.

As displayed in Figs. 5 (c, d), it can be seen that the twin well and heteroclinic orbits connected the saddle points $(\theta_1, 0)$ and $(\theta_2, 0)$. There are the periodic trajectories around the equilibrium centers $(\theta_3, 0)$ and $(\theta_4, 0)$ for Hamilton value of $0 < H < 0.5$. It is found that there are the small periodic solution denoted by the dashed line, the heteroclinic orbits marked with the dash-dotted lines and the big trajectories represented with the solid lines.

As depicted in Figs. 5 (e, f), it is found exhibits the double well and heteroclinic orbits connected the saddle points of $(\theta_1, 0)$ and $(-\theta_1, 0)$. Moreover, there are the small periodic trajectories around the equilibrium centers $(\theta_3, 0)$ and $(\theta_4, 0)$ for Hamilton value of $0 < H < PEN_1$. the big periodic solution correspond to Hamilton value of $PEN_1 < H < PEN_2$. It is shown that the dashed line denotes the small periodic solution, the dash-dotted line is the middle solution and the dotted lines represent the heteroclinic orbits. The solid lines indicate the big trajectories.

As plotted in Figs. 5 (g, h), it is found exhibits the double well and heteroclinic orbits connected the saddle points of $(\theta_1, 0)$ and $(-\theta_1, 0)$. The periodic trajectories around the equilibrium center $(\theta_2, 0)$ Hamilton value of $0 < H < PEN_2$. The big rotation periodic solution correspond to Hamilton value of $PEN_2 < H$. Then we observe that the dashed line is the small periodic solution, the dash-dotted lines represent the heteroclinic orbits and the solid lines indicate the big trajectories.

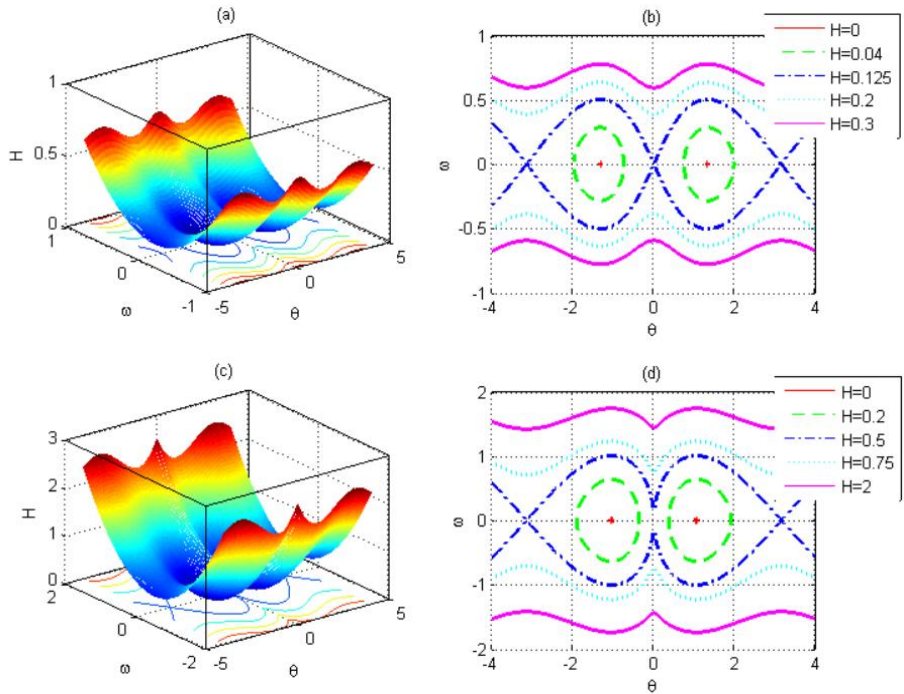

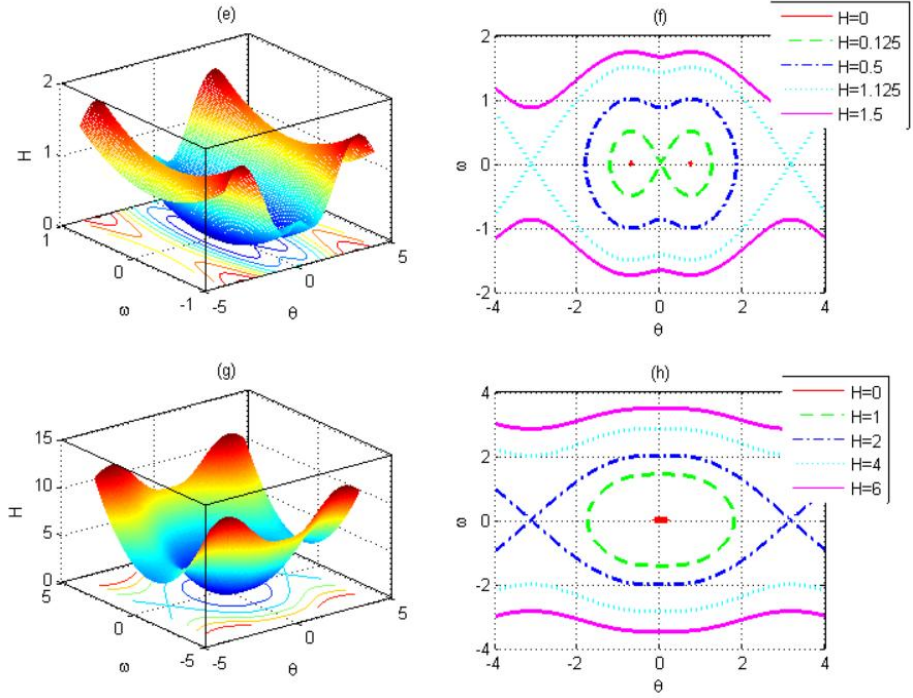

Fig. 5 Hamilton energy function surfaces and the phase trajectories for system (11) with parameter setting of $\kappa=1$ and $\gamma=0$. (a) The curved energy function $H$ surface in three dimensional $\theta-\omega-H$ parameter space. (b) The phase portrait on the two dimensional $\theta-\omega$ plane with the different value of energy $H$ for $\alpha=0.5, \beta=1$. The dashed line denotes the small periodic solution. The dash-dotted lines represent the heteroclinic orbits. The solid lines indicate the big trajectories. (c) surface and (d) phase portrait for $\alpha=1, \beta=1$. The dashed line denotes the small periodic solution. The dash-dotted lines represent the heteroclinic orbits. The solid lines indicate the big trajectories. (e) Surface and (f) phase portrait for $\alpha=1.5, \beta=1$. The dashed line denotes the small periodic solution. The dash-dotted line is the middle solution. The dotted lines represent the heteroclinic orbits. The solid lines indicate the big trajectories. (g) Surface and (h) phase portrait for $\alpha=2, \beta=1$. The dashed line denotes the small periodic solution. The dash-dotted lines represent the heteroclinic orbits. The solid lines indicate the big trajectories.

## 4. Bifurcation analysis of equilibrium for the conservative system

### 4.1 Equilibrium bifurcation

Letting nonlinear elastic moment of force equal to zero, the equilibrium angular position sets $E$ is defined as following form [33]

$$E=\{(\theta,\alpha,\beta)\,|\,M_s(\theta,\alpha,\beta)=0\} \quad (15)$$

herein those three $\theta$ is real number, $\alpha$ and $\beta$ are positive real number.

By solving the Eq. (15), the closed form solution of the equilibrium points $S_i$ of the included angle are given as follows form

$$S_i=(\theta_i,0) \quad (16)$$

where $i=1, 2, 3, 4$.

Then, the analytical expressions of the static equilibrium angle $\theta_i$ ($i=1, 2, 3, 4$) can be expressed as following formula

$$\begin{cases} \theta_1 = 2n\pi \\ \theta_2 = (2n+1)\pi \\ \theta_3 = 2n\pi + \arccos\left(\dfrac{(\alpha^2+\beta^2)(1+\gamma)^2 - 1}{2\alpha\beta(1+\gamma)^2}\right) \\ \theta_4 = 2n\pi - \arccos\left(\dfrac{(\alpha^2+\beta^2)(1+\gamma)^2 - 1}{2\alpha\beta(1+\gamma)^2}\right) \end{cases} \quad (17)$$

where $n \in Z$ and $Z$ is the integer. The angle $\theta_i$ ($i = 1, 2, 3, 4$) are the abscissas of the points of intersection of the moment of forces curves with the $x$ − axis. Therefore, there are infinite equilibrium solutions for the robot system (11).

As shown in Figs. 6(a, c), the equilibrium surfaces of parameter $\theta - \alpha - \beta$ space are given to show the bifurcation behavior for $\gamma = 0.5$ and $\gamma = 0$ respectively. In Fig. 6(b), the supercritical and sub-critical pitchfork bifurcations take place at the line $B_1$ and $B_2$ respectively. In Fig. 6(d), the subcritical bifurcation occurs at the point $(\theta, \alpha) = (0, 1)$.

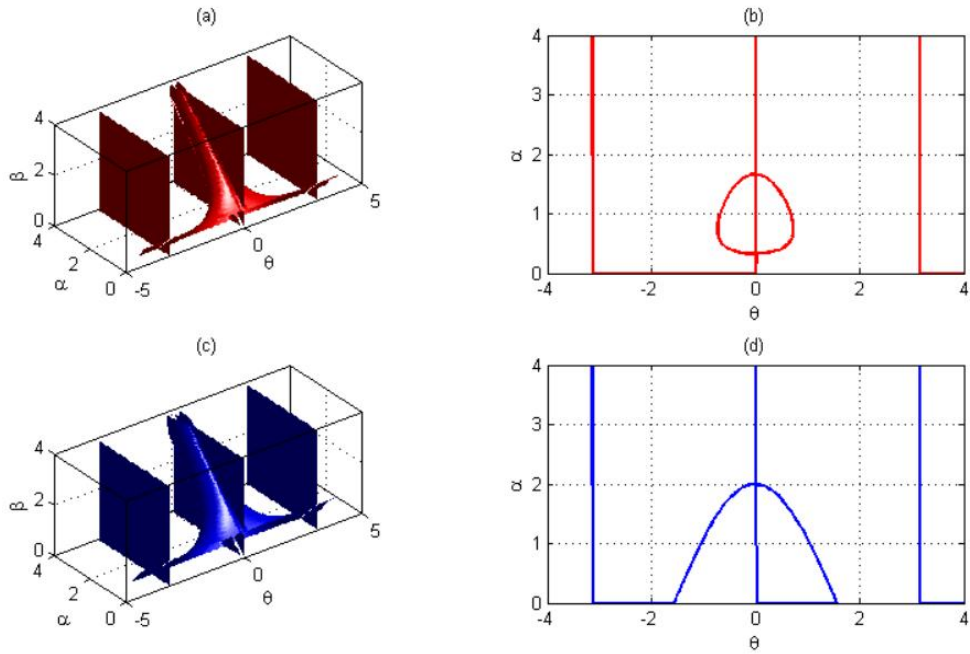

Fig. 6 Equilibrium surface parameter in tridimensional $\theta - \alpha - \beta$ space and bifurcation diagram on bidimensional $\theta - \alpha$ plane for Eq. (15). (a) Equilibrium surface and (b) bifurcation curve for $\gamma = 0.5$. (c) Equilibrium surface and (d) bifurcation curves for $\gamma = 0$.

### 4.2 Nonlinear stiffness

For fixed value of the geometrical parameters, the gradient of nonlinear moment $K = dM_s/d\theta$ is the nonlinear stiffness and can be defined as form

$$K = \left(\alpha\beta + \gamma - \frac{\alpha\beta}{(\alpha^2 + \beta^2 - 2\alpha\beta\cos\theta)^{1/2}}\right)\cos\theta + \frac{(\alpha\beta\sin\theta)^2}{(\alpha^2 + \beta^2 - 2\alpha\beta\cos\theta)^{3/2}} \quad (18)$$

here $K$ is the complex stiffness function with both fractal and radical non-linearity. It is also found that the stiffness is nonlinear function of the angle variable $\theta$ and the geometrical parameters $\alpha$, $\beta$ and the gravitational coefficient $\gamma$.

As shown in Fig. 7(a, c), the nonlinear stiffness $K$ surfaces are given in the parameter space $\theta - \alpha - \beta$. In Fig. 7(b) the positive, negative and zero stiffness characteristic are obtained to design the flapping mechanism of wing robot system. Nonlinear stiffness and hardness and soften formula at the left and right side of the equilibrium $(\theta_3, 0)$ or at the right and left side of the equilibrium $(\theta_4, 0)$ respectively. Therefore, the soft stiffness characteristic in angular region $\theta_3 < \theta < \pi$ or $-\pi < \theta < \theta_4$ are useful in engineering application.

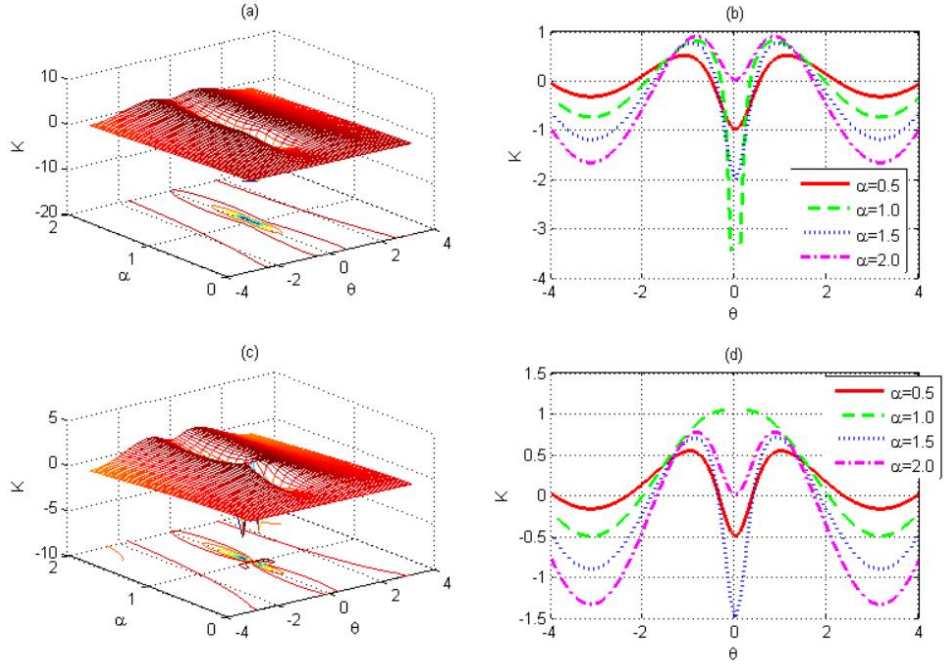

Fig. 7 The nonlinear analysis of dimensionless elastic stiffness $K$. (a) The curved stiffness surface $K$ in the three dimensional $\theta - \alpha - K$ space and (b) the nonlinear stiffness $K$ curves on the two dimensional $\theta - \alpha - K$ plane for different values of geometrical parameter $\alpha$. for $\gamma = 0$. (c) The stiffness surface $K$ and (d) the stiffness $K$ curves for different values of $\alpha$ and $\gamma = 0.5$.

By letting the stiffness of Eq. (17) equal to zero, the bifurcation set in the expression of the implicit function can be defined and obtained as follows

$$B_0 = \{(\alpha, \beta, \gamma) \,|\, K = 0\} \tag{19}$$

herein $\alpha \geq 0$, $\beta \geq 0$, $\gamma \in R$ and $R$ is real number. The angular solution $\theta = \theta(\alpha, \beta, \gamma)$ can not be written in a closed form due to the essential mathematical difficulties.

As shown in Fig. 8(a), the zero stiffness surface is obtain with help of the computational method. It is found that the zero stiffness relationship is a curved surface in the three dimensional $\theta - \alpha - \gamma$ parameter space. In Fig. 8(b), the zero stiffness curves are plotted on the bidimensional $\theta - \alpha$ parameter plane for the different value of gravitational ratio $\gamma = 0, 0.5, 1.0$ and $1.5$. It is shown that the curves demonstrate the multivalued characteristic.

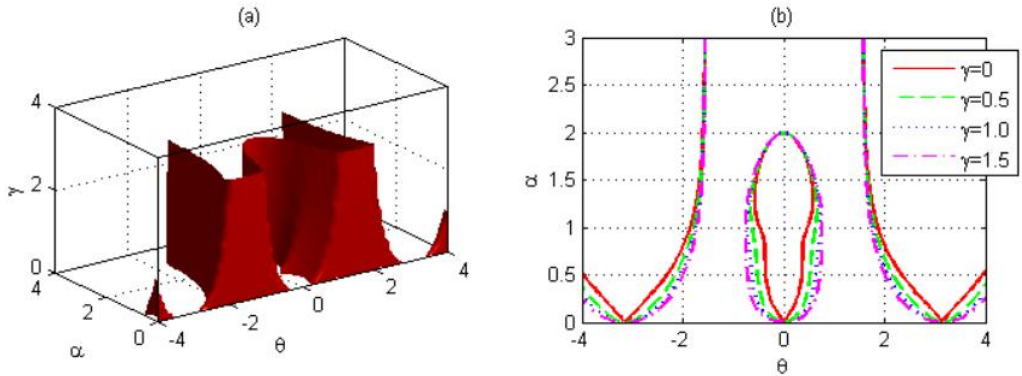

Fig. 8 The surface and the curves for the zero stiffness $B_0$. (a) The zero surface in the three dimension $\theta - \alpha - \gamma$ parameter space and (b) the nonlinear curves on the two dimensional parameter $\theta - \alpha$ plane for the different valve of the gravitational ratio $\gamma = 0, 0.5, 1.0$ and $1.5$.

In order to investigate the free vitiation behaviors near the angular equilibrium $\theta_i$, the nonlinear stiffness $K$ at equilibrium points of angle $\theta_i$ are computed and obtained as follows

$$K_i = (\alpha\beta + \gamma)\cos\theta_i - \frac{\alpha\beta\cos\theta_i}{(\alpha^2 + \beta^2 - 2\alpha\beta\cos\theta_i)^{1/2}} + \frac{(\alpha\beta\sin\theta_i)^2}{(\alpha^2 + \beta^2 - 2\alpha\beta\cos\theta_i)^{3/2}} \quad (20)$$

here $\theta_i$ ($i = 1, 2, 3, 4$) is the equilibrium angle of Eq. (16). The stiffness $K$ have the relationship with the three parameters $\alpha, \beta$ and $\gamma$.

On the one hand, for the equilibrium angle $\theta_i$ ($i = 1, 2$), submitting those $\theta_i$ into the Eq. (19) and leading to analytical formula as follows

$$\begin{cases} K_1 = \left((\alpha\beta + \gamma) - \dfrac{\alpha\beta}{|\alpha - \beta|}\right) \\ K_2 = -\left((\alpha\beta + \gamma) - \dfrac{\alpha\beta}{|\alpha + \beta|}\right) \end{cases} \quad (21)$$

where $i = 1, 2$. It is found that the stiffness $K_1$ are the negative stiffness and the corresponding equilibrium $S_1$ is saddle. The stiffness $K_2$ is the negative stiffness for region IV and the corresponding equilibrium $S_2$ is saddle. The stiffness $K_2$ is positive stiffness for regions I, II, III.

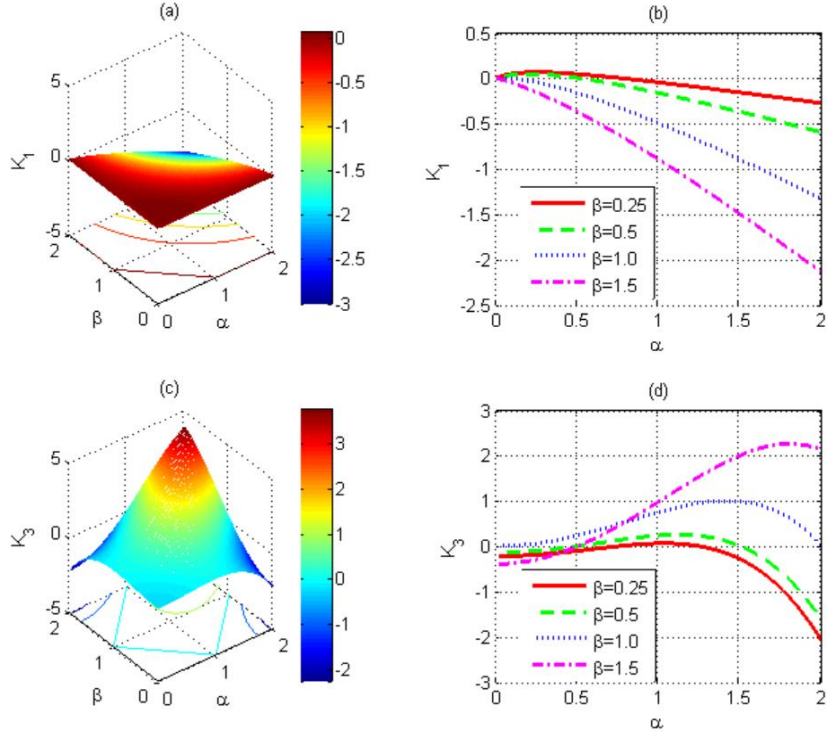

Fig. 9 The nonlinear stiffness $K$ of Eq. (19) at the equilibrium angles $\theta_{2,3}$. (a) Stiffness surface in tridimensional parameter $\alpha - \beta - K$ space for Eq. (20) and (b) curves on bidimensional parameter $\alpha - K$ plane for different value of parameter $\beta = 0.25, 0.5, 1.0, 1.5$. (c) Stiffness surface in $\alpha - \beta - K$ space for Eq. (20) and (d) curves on $\alpha - K$ plane for different geometrical ratio $\beta = 0.25, 0.5, 1.0, 1.5$.

On the other hand, we submitting equilibrium angle $\theta_3$ into the Eq. (19) and leading to closed form expression of stiffness $K_3$ as following form

$$K_3 = \frac{\left(\alpha\beta\left(1 - \left(\dfrac{(\alpha^2 + \beta^2)(1+\gamma)^2 - 1}{2\alpha\beta(1+\gamma)^2}\right)^2\right)\right)}{\left(\alpha^2 + \beta^2 - \dfrac{(\alpha^2 + \beta^2)(1+\gamma)^2 - 1}{(1+\gamma)^2}\right)^{3/2}} \quad (22)$$

where $K_3 > 0$. It is easy to know that the stiffness at $\theta_4$ is $K_4 = K_3$ due to the symmetry. It is found that the stiffness $K_{3,4}$ is the positive stiffness for regions IV and the corresponding equilibrium $S_{3,4}$ are the singular points of the centers.

The nonlinear stiffness $K_2 = K(\alpha, \beta)$ curved surface is plotted in Fig. 9(a) in three dimensional space parameters for $\gamma = 0$. It is found that the smaller geometrical parameters $\alpha$ and $\beta$ lead to the larger stiffness $K_2$. A clearer trend in stiffness variation is shown in the Fig. 9(b). It is shown that the stiffness values decrease when the different values of the geometrical parameter increase of $\beta = 0.25, 0.5, 1.0, 1.5$.

The $K_{3,4} = K(\alpha, \beta)$ surface is plotted in Fig. 9(a) in three dimensional space parameters for $\gamma = 0$. It is found that the bigger geometrical parameters $\alpha$ and $\beta$ corresponding to the larger stiffness $K$. A clearer trend in stiffness variation is shown in the Fig. 9(b). It is shown that the stiffness firstly increase and then decrease with the different value of parameter $\beta = 0.25, 0.5, 1.0, 1.5$.

### 4.3 Bifurcation set analysis

In order to reveal parameter condition of transaction mechanism of the multiple stability, the bifurcation sets are defined as follows

$$B = \{(\alpha, \beta, \gamma) | F_s = 0, K = 0\} \tag{23}$$

where $\alpha, \beta, \gamma \geq 0$ is the positive real number.

By combining Eq (13) and Eq. (17), the analytical expression of the bifurcation set surface $B = B_1 \cup B_2$ are computed and written as following

$$\begin{cases} B_1 = B_{\theta=2n\pi} = \left\{(\alpha, \beta, \gamma) | \alpha\beta - \dfrac{\alpha\beta}{|\alpha - \beta|} + \gamma = 0\right\} \\ B_2 = B_{\theta=(2n+1)\pi} = \left\{(\alpha, \beta, \gamma) | \alpha\beta - \dfrac{\alpha\beta}{|\alpha + \beta|} - \gamma = 0\right\} \end{cases} \tag{24}$$

where $n$ is the positive integral number. It is noted that $B$ has the relationship with three system parameters of $\alpha, \beta$ and $\gamma$. Based on Eq. (24), it is convenient to investigate the transition characteristic and classify the dynamic response.

As presented in Fig. 10, the bifurcation sets are plotted in the tridimensional parameter space $\alpha - \beta - \gamma$ and the bidimansional parameter plane $\alpha - \beta$. In Figs. 10 (a, b), the curved surfaces of bifurcation set are used to divide space into the region of single well and double well cases. As shown in Figs. 10 (c, d), the bifurcation $\alpha - \beta$ plane are divided into four regions, at which $I, II$ and $III$ for single stable of soft characteristic of Duffing system and at region $IV$ for double stable of double well of Duffing system. The correspond phase portraits on the bidimansional parameter $\alpha - \beta$ plane are given in Fig. 5. The results show that the system parameters located in the region $IV$ which can be beneficial to the efficiently flapping of the dipteran robot.

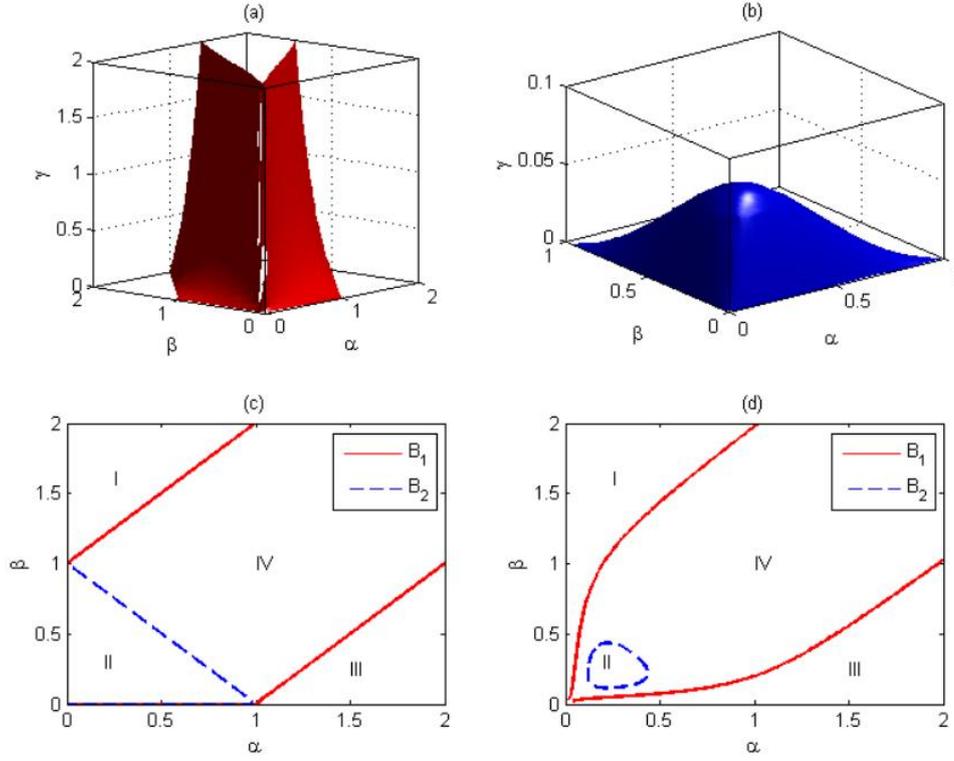

Fig. 10 The nonlinear bifurcation diagram. (a) The curved surface in the $\alpha-\beta-\gamma$ space for the bifurcation set $B_1$ and (b) bifurcation set $B_2$. (c) The bifurcation curves on $\alpha-\beta$ plane of $B_1$ for $\gamma=0$. (d)The bifurcation curves $B_2$ for $\gamma=0.05$.

### 4.4 Stability of equilibrium

For the conservative robot system (11), the Jacobian matrix at the singular points of the equilibria $S_i$ are obtained as follows

$$J_i = \begin{pmatrix} a_{11} & a_{12} \\ a_{21} & a_{22} \end{pmatrix} \tag{25}$$

where $a_{11}=0, a_{12}=1, a_{21}=-K(\theta_i), a_{22}=0$ and $i=1,2,3,4$.

It is easily noted that $J_i$ is a $2\times 2$ matrix as well as the characteristic polynomial and the eigenvalue can be computed quickly. Therefore, the characteristic equation of the matrix (25) are defined as following form

$$\det(\lambda I - J_i) = \lambda^2 - p\lambda + q = 0 \tag{26}$$

where $I$ is an identity matrix, $\lambda$ is called the eigenvalue of Jacobian $J_i$, $p=\text{tr}(J_i)=a_{11}+a_{22}, q=\det(J_i)=a_{11}a_{22}-a_{12}a_{21}$ and $i=1,2,3,4$.

By using quadratic formula, the eigenvalues of the Jacobian $J_i$ determinant of Eq. (26) are obtained as following formula

$$\lambda_{1,2}(\theta_i) = \frac{1}{2}(p \pm \sqrt{\Delta}) = \pm\sqrt{-K(\theta_i)} \tag{27}$$

where $\Delta = p^2 - 4q$ is discriminant and $i=1,2,3,4$.

As shown in Fig. 11(a), the we obtain and classify the character of various singular points. The entire trajectories structure and the corresponding stability of singular points are studied. Furthermore, we divide the parameter $p-q$ plane into five regions which characterized different singular points.

The singular classification are given in Fig. 11(a). It is found that the character of the conservative system (11) depends on the eigenvalue of $J_i$ and on the parameters $p$ and $q$ according to the Eq. (26). In Fig. 11(b), while $q<0$ corresponds to the saddle points. In Fig. 11(c), when $\Delta>0$, $p>0$ and $p<0$ correspond to unstable and stable nodes respectively. In Fig. 11(d), for $\Delta<0$, $p>0$ and $p<0$ correspond to unstable and stable foci respectively. The curve $\Delta=0$ of repeated eigenvalues separated nodes from focus points, corresponds to degenerate node. While $p=0$ and $q>0$, which separates the unstable from the stable focus point, corresponds to centers.

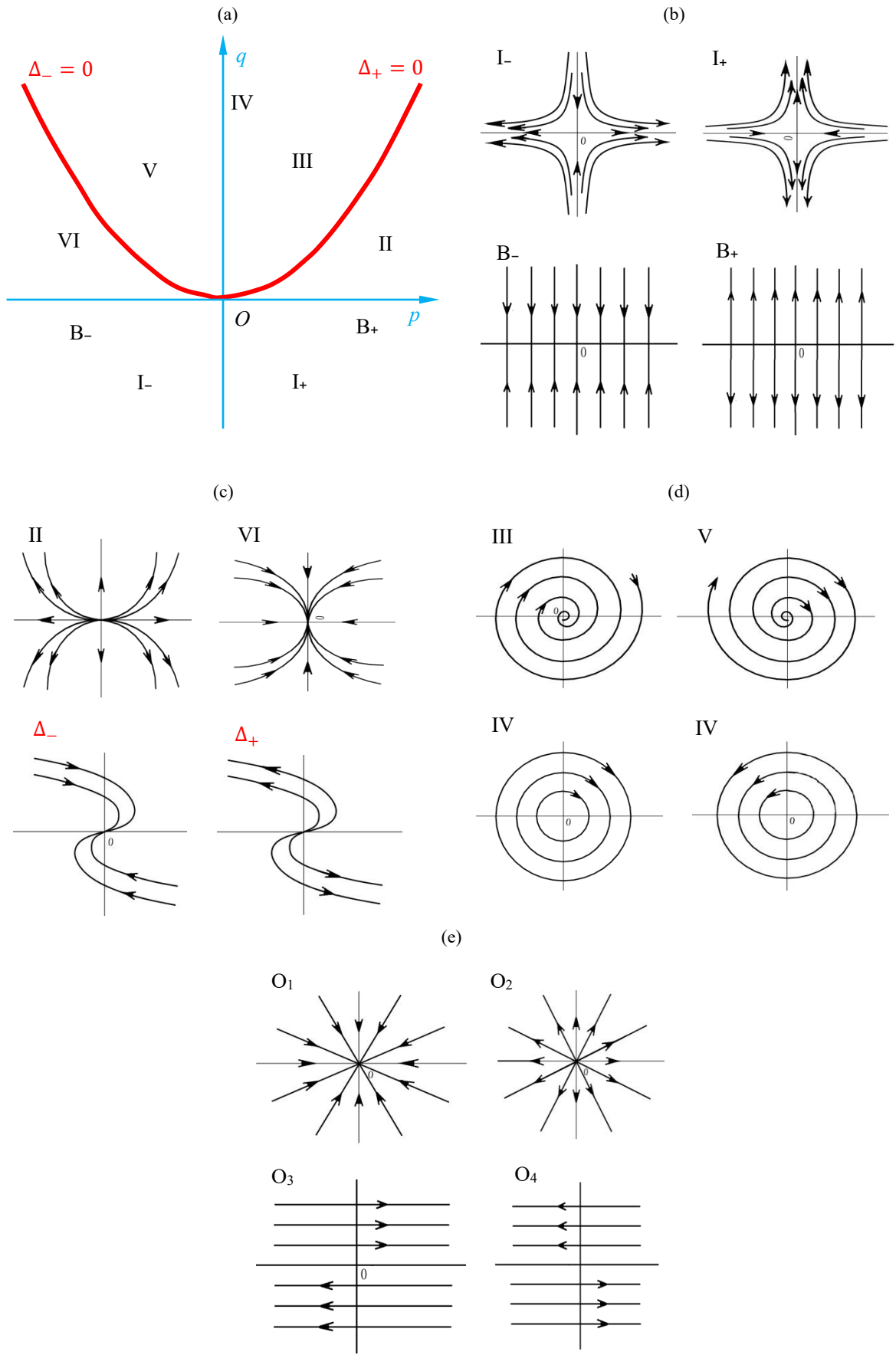

Fig. 11 Singularity points. (a) Classification on the parameters $p-q$ plane. (b) The saddles $I_\pm$, stable and unstable singular lines $B_\mp$. (c) The unstable node II and stable node IV and unstable and stable degenerate nodes $\Delta_\pm$ for $\Delta = 0$. (d) The stable and unstable foci for III and V respectively, centers for IV. (e) The trajectories $O_{3,4}$ for $\Delta = 0$ and $O_{3,4}$ for $\text{Det}(J_i) = 0$.

For equilibrium points $S_i$ ($i=1,2$) the corresponding two eigenvalues of positive and negative real roots are obtained as

$$\lambda_{1,2}(\theta_1) = \pm\sqrt{-K(\theta_1)} \tag{28}$$

where $i = 1, 2$. The geometrical parameters satisfied the condition of the point $(\alpha, \beta) \in$ IV shown in Fig. 10(c) and the corresponding stiffness $K(\theta_i) < 0$. Therefore, it is found that the equilibrium $S_i$ $(i = 1, 2)$ is a saddle point. The phase ports diagram as shown in Fig.11(b). Whereas, when $(\alpha, \beta) \in$ I, II, III, the equilibrium point $S_1$ $(i = 1, 2)$ is a saddle and the point $S_2$ is a center (See Fig. 11(d)).

Clearly, for the a saddle points of the equilibrium $S_i$ $(i = 1, 2)$, the mode matrix of the eigenvectors corresponding to eigenvalue Eq. (28) are obtained as follows

$$\begin{pmatrix} 1 & 1 \\ \sqrt{-K(\theta_1)} & -\sqrt{-K(\theta_1)} \end{pmatrix} \tag{29}$$

where $K(\theta_1) < 0$.

Furthermore, for the equilibrium points $(\theta_{1,2}, 0)$, the corresponding eigenvalues of a pair of imaginary roots are obtained as following

$$\lambda_{1,2}(\theta_i) = \pm j\sqrt{K(\theta_i)} \tag{30}$$

where $i = 3, 4$ and $j = \sqrt{-1}$ is imagery unit, parameter in the region $(\alpha, \beta) \in$ VI and nonlinear stiffness $K(\theta_{3,4}) > 0$. Hence, it is found that the equilibrium $(\theta_{3,4}, 0)$ is a center (see Fig. 11(d)). The phase ports diagram as shown in Fig.11(d).

Thus, for the centers point of equilibrium $S_i$ $(i = 1, 2)$, the feature matrices of the complex eigenvectors corresponds to the eigenvalue Eq. (30) are given by

$$\begin{pmatrix} 1 & 1 \\ j\sqrt{-K(\theta_1)} & -j\sqrt{-K(\theta_1)} \end{pmatrix} \tag{31}$$

where $j = \sqrt{-1}$ is imagery unit and $K(\theta_i) > 0$ $(i = 3, 4)$.

## 5. Flapping dynamic response

### 5.1 Periodic solution of the conservative vibration system

According to the discussion above, it is known that periodic vibration around the center exist with the eigenvalue Eq. (30). Therefore, we investigate those periodic motions, and the exact solution of Eq. (21) can be obtained by separated variables. The result of the period is integrated and obtained as [34]

$$T = \int_{\theta_0}^{\theta} \frac{d\theta}{\sqrt{2H - \left(\sqrt{\alpha^2 + \beta^2 - 2\alpha\beta\cos\theta} - 1\right)^2 - \gamma(1 - \cos\theta)}} \tag{32}$$

where $T$ are the dimensional periodic time and $\theta_0$ are the initial angle displacement and $H > 0$ is the total energy. In other words, we assumed that $T = 0$ when $\theta = \theta_0$.

For the nondimensional parameter condition $(\alpha, \beta) \in$ I, II, III (see Fig. 10), the Jacobi elliptical function of the sine amplitude is defined based on Eq. (32) and obtained as follows

$$\theta = \theta_0 \text{sn}(T, k) \tag{33}$$

where $T > 0$ is the time and $k = 2H/(1 - |\alpha - \beta|)^2 \in (0, 1)$ is called the modulus of the elliptic functions.

For the nondimensional parameter condition $(\alpha, \beta) \in$ IV (see Fig. 10), the Jacobi elliptical function of the cosine amplitude is defined based on Eq. (32) and obtained as follows

$$\theta = \theta_0 \text{cn}(T, k) \tag{34}$$

where $T > 0$ is the time and $k = 2H/(1 - |\alpha - \beta|)^2 \in (0, 1)$ is the elliptic modulus.

For the nondimensional parameter condition $(\alpha, \beta) \in$ IV (see Fig. 10), the Jacobi elliptical function of the delta amplitude is defined based on Eq. (32) and obtained as follows

$$\theta = \theta_0 \mathrm{dn}(T, k) \tag{35}$$

where $T > 0$ is the time and $k = (1 - |\alpha - \beta|)^2/(2H) \in (0, 1)$ is the elliptic modulus.

As shown in Fig. 11, the periodic response solutions for double well system (10) are plotted for the different initial value $(\theta_0, \omega_0)$. In Fig. 11,

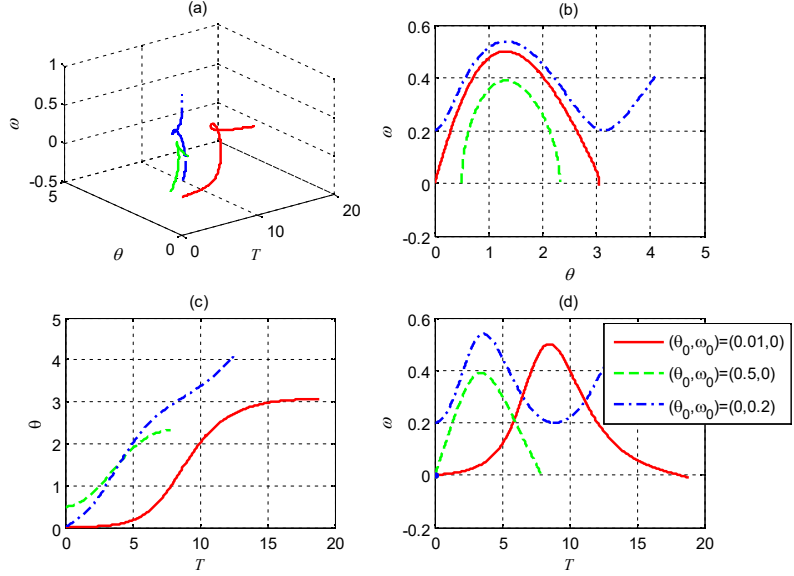

Fig. 12 The nonlinear vibration response with $\kappa = 1$ and $\gamma = 0$ for initial conditions $(\theta_0, \omega_0) = (0.01, 0)$, $(\theta_0, \omega_0) = (0.5, 0)$ and $(\theta_0, \omega_0) = (0, 0.2)$. Solid line denotes the hetero-clinic $\theta_{het}$ orbit. Dashed line represents the $\theta_{dn}$ function. Dash-dotted line indicates the $\theta_{cn}$ function. (a) Three dimensional trajectories and (b) phase portrait. (c) The time histories of angle and (d) angular velocity.

For the free vibration of robot system (11), the natural frequency $\Omega_n$ at the equilibrium angle $\theta_{1,2}$ can be expressed as follows

$$\Omega_n = \sqrt{\frac{K}{\kappa}} \tag{36}$$

where $\Omega_n > 0$ is the free vibration frequency at equilibrium points $S_{1,2}$.

As shown in Fig. 12(a), the stiffness surface is obtained. In Fig. 12(b), the curves of intrinsic frequency $\Omega_n$ are plotted for different value of geometrical parameter $\beta = 0.25, 0.5, 1.0$ and $1.5$.

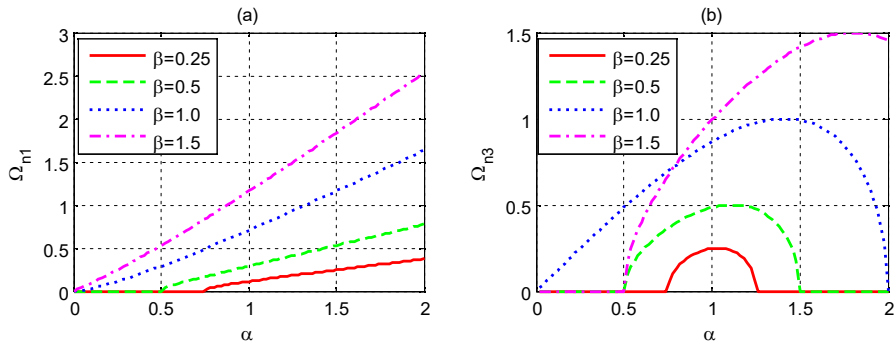

Fig. 13 The free vibration characteristics at the equilibrium points $S_2$ and $S_3$ on the parameter $\alpha - \Omega_n$ plane. (a) The free vibration frequency $\Omega_{n1}$ and (b) frequency $\Omega_{n3}$ curves for different value $\beta$ geometrical parameter $\beta = 0.25, 0.5, 1.0$ and $1.5$.

For gravitational parameters $\gamma = 0$, making using of Hamilton formula (21), the period can be integrated and obtained as follows

$$\int_{T_{ini}}^{T_{fin}} dT = \int_{\theta_{ini}}^{\theta_{fin}} \frac{d\theta}{\sqrt{2H - \left(\sqrt{\alpha^2 + \beta^2 - 2\alpha\beta\cos\theta} - 1\right)^2 - \gamma(1 - \cos\theta)}} \tag{37}$$

where $0 < T_{ini} < T_{fin}$ are the dimensional time and $0 < \theta_{ini} < \theta_{fin}$ are the angle displacement and $H > 0$ is the total energy.

Based on Hamilton function Eq. (14), the angular displacement of the intersect points between the closed periodic trajectory and the abscissas axis are solved and obtained as follows

$$\begin{cases} \theta_{ini} = \arccos\left(\frac{\alpha^2 + \beta^2 - (\sqrt{2H} + 1)^2}{2\alpha\beta}\right) \\ \theta_{fin} = \arccos\left(\frac{\alpha^2 + \beta^2 - (\sqrt{2H} - 1)^2}{2\alpha\beta}\right) \end{cases} \quad (38)$$

where $\theta_{ini}$ is the initial angle and $\theta_{fin}$ is the finished angle, as well as $0 < H < H_1$ and $H_1 = H(\theta = 0) = 0.5(|\alpha - \beta| - 1)^2$ is a energy value at the angle $\theta = 0$.

For the free oscillation, the amplitude of the angle is the absolute value of the difference between the maximum $\theta_{fin}$ and minimum $\theta_{ini}$ values defined as

$$A_f = \frac{1}{2}|\theta_{fin} - \theta_{ini}| \quad (39)$$

where $\theta_{ini}$ is the initial angle and $\theta_{fin}$ denote the final angle.

The period of free vibration is the difference between the maximum $T_{fin}$ and minimum $T_{ini}$ values are defined as follows

$$T = T_{fin} - T_{ini} \quad (40)$$

here $T_{ini}$ is the initial time and $T_{fin}$ denote the final time.

The amplitude of the oscillation frequency is equal to $2\pi$ divided by the period $T$, defined as follows

$$\Omega = \frac{2\pi}{T_{fin} - T_{ini}} \quad (41)$$

where $T_{ini}$ is the initial time and $T_{fin}$ denote the final time.

If parameter located in regions I, II and III (see Fig. 10). The magnitude of the amplitude depends on the magnitude of the total energy. For small finite oscillation around the center $S_1$, meanwhile the Hamilton energy satisfied with the region

$$\begin{cases} AF_1 = \{(\Omega_n, A_f) \mid 0 < H < H_2\} \\ AF_2 = \{(\Omega_n, A_f) \mid H_2 < H\} \end{cases} \quad (42)$$

where $H_2 = H(\theta_2 = \pi) = 0.5(|\alpha + \beta| - 1)^2$. The amplitude frequency curve is sketched plotted in Fig. 14(a) and denoted by the solid line of $H < H_2$ and dotted line for $H > H_2$. As a result, it is observed that the amplitude frequency displays the soft Duffing properties of left bending response.

If parameter located in regions IV (see Fig. 10). For small finite oscillation around the center $S_{3,4}$, when the Hamilton energy $H$ satisfied the region

$$\begin{cases} AF_3 = \{(\Omega_n, A_f) \mid 0 < H < H_2\} \\ AF_4 = \{(\Omega_n, A_f) \mid H_1 < H < H_2\} \\ AF_5 = \{(\Omega_n, A_f) \mid H_2 < H\} \end{cases} \quad (43)$$

where $H_1 = H(\theta_1 = 0) = 0.5(|\alpha - \beta| - 1)^2$. The corresponding amplitude frequency curve as shown in Fig. 14(b) denoted by the dashed line for $H < H_1$ and dashed line for $H_1 < H < H_2$. Moreover, the amplitude frequency curve of $H > H_1$ as shown in Fig. 14(b) denoted by the dotted line. As a result, it reveals that the amplitude frequency exhibits both the soften of inter well and the harden of outer well characteristics.

(a)         (b)

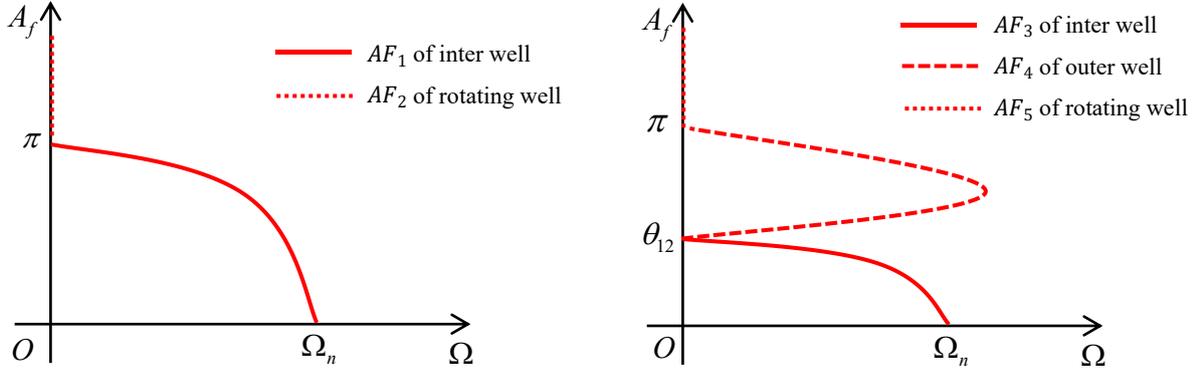

Fig. 14 The frequency response curves for system (10). (a)Amplitude frequency curves, solid line $AF_1$ represents the small periodic orbit and the dotted line $AF_2$ denotes the big periodic trajectories for the double well corresponding parameter region I, II, III. (b)Amplitude frequency curves, solid line $AF_3$ and dashed line $AF_4$ represent the inter and outer response of separatriex trajectory for the double well in parameter region IV, and dotted line $AF_5$ denotes the bigger periodic response.

**5.2 The periodic response of the perturbed robot system**

The approximate moment of force (13) can be obtained by using of the topological method and the dimensionless governing equation of the perturbed dipteran flight robot system becomes

$$\kappa\ddot{\theta} + 2\xi\omega + \Omega_n^2\left(\theta + \varepsilon\theta^3\right) = M_0\sin(\Omega_0 T + \varphi) \tag{44}$$

where $\omega_0 = \pi\sqrt{K}$ is the natural frequency, $\varepsilon = -K/\pi$ is the nonlinear coefficient and $\varphi$ is the phase. The multivaluedness of the response curves due to the nonlinearity $\varepsilon$ has a significance for that it lead to jump phenomena.

For flapping robot, the vibration of the mass about the equilibrium angle $\theta_2$ or $\theta_{3,4}$, which are always happening in actual flight. Therefore, the approximate solution can be obtained by the Harmonic Balance Method (HBM) [30]. The general solution of the perturbed system (45) can be written as follow

$$\theta = A_p \sin(\Omega_0 T) \tag{45}$$

Substituting the solution (46) into Eq.(45) and equating the coefficients of $\sin(\Omega_0 T)$ and $\cos(\Omega_0 T)$ on both sides, we obtain the relationship as follows

$$\begin{cases} A_p(1-\kappa s^2) + \dfrac{3}{4}\varepsilon A_p^3 = \dfrac{M_0}{\Omega_0}\cos\varphi \\ 2\xi s A_p = \dfrac{M_0}{\Omega_0}\sin\varphi \end{cases} \tag{46}$$

where $\Omega_n^2 = K\pi^2$ and $s = \Omega_0/\Omega_n$ is the frequency ratio. We note that the excitation, nonlinear, damping and inertia terms appear in Eq. (47)

Eliminating phase $\varphi$ from Eqs. (47) by computing square and adding, relationship equation of the amplitude frequency is obtained as following

$$\left(\left(1-\kappa s^2 + 0.75\varepsilon A_p^3\right)^2 + (2\xi s)^2\right)A_p^2 = B^2 \tag{47}$$

where $s = \Omega_0/\Omega_n$ and $B = M_0/\Omega_n$. The mulivalueness and jump phenomena of Eq. (48) can be explained by the catastrophe theory [Nayphy 1995].

Eq. (47) is an implicit equation function for the amplitude of response $A_p$ and The amplitude $A_p$ as a function of the system quantities $s$, $\kappa$, $B$, $\varepsilon$ and $\xi$, gives

$$A_p = \frac{B^2}{\sqrt{\left(1-\kappa s^2 + 0.75\varepsilon A_p^2\right)^2 + (2\xi s)^2}} \tag{48}$$

where $A_p$ is the response amplitude of the perturbed system (45).

Through the division of Eq. (46), the phase frequency relationship for the system (44) is given in the following form

$$\varphi = \arctan\left(\frac{2\xi s}{1-\kappa s^2 + 0.75\varepsilon A_p^2}\right) \quad (49)$$

where $\varphi$ is the phase angle of the perturbed system (44).

As shown in Fig. 14, the influence of nonlinear coefficient $\varepsilon$ on the amplitude frequency surface are plotted in three parameter $s-\xi-A_p$ space with $B=1$ and $\kappa=1$. It is found that the nonlinear parameter $\varepsilon$ bends the amplitude frequency surface away from the linear $\varepsilon=0$ to the left for the soft spring $\varepsilon<0$. It is noted that some of amplitude frequency surface are multiple values and while others are single value depending on the nonlinear parameter value of $\varepsilon$. The multivaluedness is responsible for the jump phenomenon. In Fig. 15(b), the peak amplitude is infinite in the absence of damping ratio ($\xi=0$). The locus of peak amplitudes is a parabola formula $s=1+0.75\varepsilon a^2$ that is usually named the backbone curve.

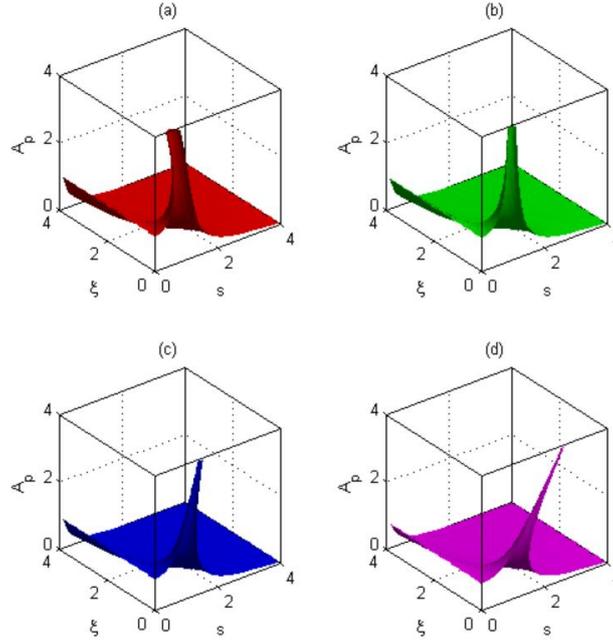

Fig. 15 The effect of stiffness parameter $\varepsilon$ on the amplitude frequency surface for the primary resonance of the perturbed system (45). (a) $\varepsilon=-0.05$, (b) $\varepsilon=0$, (c) $\varepsilon=0.05$, (d) $\varepsilon=0.5$.

### 5.3 Chaotic vibration of forced vibration robot system

#### 5.3.1 Approximated method of the double well Duffing

Based on the topological equal method, the two dimensional first-order equations of unperturbed dipteran flight robot system with homo-clinic orbit is obtained, and one has

$$\begin{cases} \dot{\theta} = \omega \\ \kappa\dot{\omega} = -2\xi_0\omega - \theta(\theta-\theta_3)(\theta-\theta_4) + M_0\sin(\Omega_0 T) \end{cases} \quad (50)$$

where $\theta$ is angle and $\omega$ is the angular speed. The equilibrium angles $\theta_{3,4}$ are refer to Eq. (16).

A first integral of the system (51) of the Hamilton function for the autonomous system (51) without damping and external moment of force can be written as follows

$$H_3 = \frac{1}{2}\omega^2 + \frac{1}{4}(\theta-\theta_3)^2(\theta-\theta_4)^2 \quad (51)$$

where $\theta_{3,4}$ is equilibrium angle and defined by Eq. (16).

The Hamilton function surface with Duffing potential wells is plotted in Fig. 16(a). The undulating surface denotes the Hamilton function of the total energy level, while the curves on the phase plane $\theta-\omega$ represent total energy levels. As can be seen in Fig. 16 (b), the phase portraits with double well behaviors are plotted for the different value $H_3$ of Hamilton energy (52). It is found that there are two centers $S_{3,4}$ and one saddle $S_2$ for the conservative robot system (51).

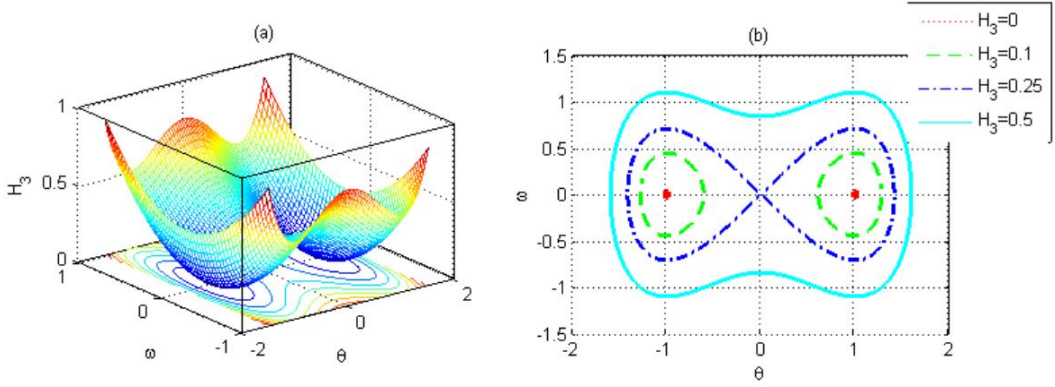

Fig. 16 Hamilton function $H_3$ surface and phase portraits for system (45). (a) Energy surface and (b) trajectories for the different values of Hamilton $H_3$.

The homo-clinic orbit function for the autonomous system (51) without the damper and the exciter are obtained in analytical form as follows

$$\begin{cases} \theta_{\text{hom}}(T) = \sqrt{2}\theta_3 \text{sech}(T) \\ \omega_{\text{hom}}(T) = -\sqrt{2}\theta_3 \text{sech}(T)\tanh(T) \end{cases} \quad (52)$$

where $T > 0$ is the dimensionless time and the equilibrium angular $\theta_3$ refer to Eq. (16).

The Melnikov function of the forced robot system (50) with the external sinusoidal excitation is computed by using the residues and obtained as following

$$M(T_0) = -\xi_0 \frac{4\theta_1^3}{3} \pm \pi M_0 \Omega_0 \frac{\sqrt{2}}{\theta_1} \text{sech}\left(\frac{\pi \Omega_0}{2\theta_1}\right) \sin(\Omega_0 T_0) \quad (53)$$

where $|\sin(\Omega T_0)| \leq 1$ and the $\cosh(*)$ is the hyperbolic cosine function.

To find the criteria for the system (51), we start setting the condition of simple zero $M(T_0) = 0$, and yield to

$$M_0 > \frac{4\theta_1^3 \xi_0}{3\sqrt{2}\pi\Omega_0} \cosh\left(\frac{\pi \Omega_0}{2\theta_1}\right) = R_1(\xi_0, \Omega_0) \quad (54)$$

which is the Melnikov zero condition $R_1 = 0$ for the chaotic vibration criteria. The $\cosh(*)$ is the hyperbolic cosine function.

In Fig. 17(a), the chaotic critical surface is given in three dimensional parameters space $\Omega_0 - \xi_0 - M_0$. The chaotic may happen when the parameters setting above this surface. In Fig. 17(b), the chaotic curves for different damping ratio $\xi_0 = 0.1, 0.2, 0.3$ and $0.4$. The chaotic behaviors will take place for non-dimensional force $F_0$ the greater than the critical lines. Fig. 17(c) shows the chaotic threshold curves of system (51) for the different external periodic excitation frequency $\Omega_0 = 0.1, 0.2, 2$ and $5$. Fig. 17(d) presents the chaotic critical curves for different external amplitude $F_0 = 0.2, 0.4, 0.6$, and $0.8$.

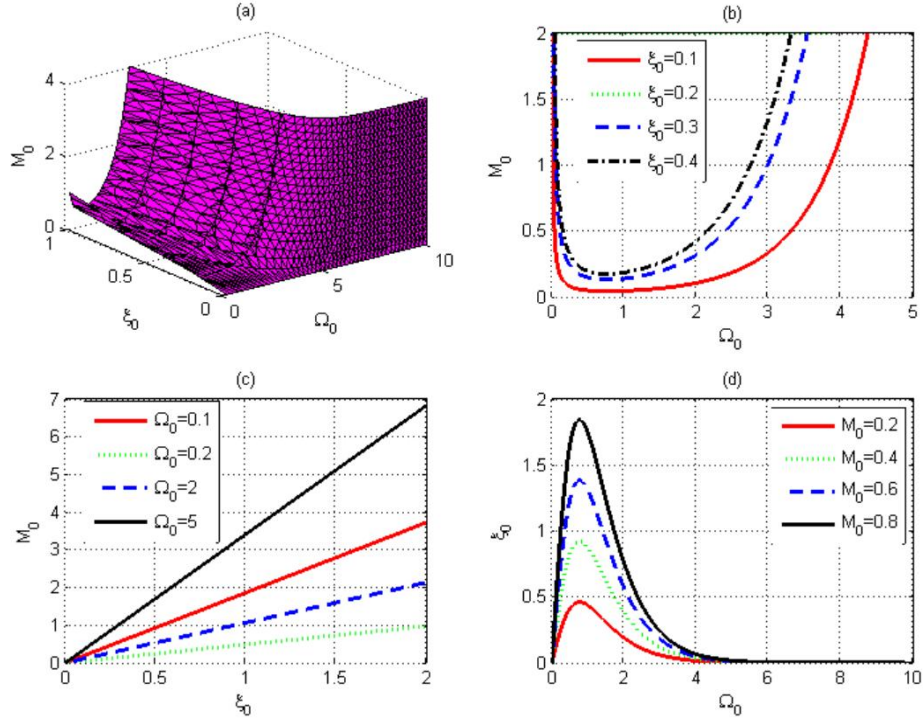

Fig. 17 Chaotic thresholds of system (45). (a) chaotic criteria surface and (b) curves for different value $\xi_0$. (c) For different value $\Omega_0$. (c) For different value $M_0$

### 5.3.2 Approximation strategy of the simple pendulum

Based on the mathematical skill of the topological equal, tor hetero-clinic orbit of the fighting robot system, the two first-order equations of unperturbed dipteran flight robot system are obtained as follows

$$\begin{cases} \dot{\theta} = \omega \\ \kappa\dot{\omega} = -2\xi_0\omega - K_1\sin\theta + M_0\sin(\Omega_0 T) \end{cases} \quad (55)$$

where $K_1 = K(\theta_1)$ are the stiffness at the equilibrium $\theta_1 = \pi$.

For the conservative system (56) without the viscous damping and forced external excitation, the Hamilton function for the autonomous system of Eq. (50) is

$$H_4 = \frac{1}{2}\omega^2 + (1-\cos\theta) \quad (56)$$

where $\theta$ is angle and $\omega$ is the angular speed.

The Hamilton function surface with periodic potential wells is plotted in Fig. 18(a). The undulating surface represents the Hamilton function of the total energy level, while the curves on the variation of $(\theta, \omega)$ the phase plane represent total energy levels. With the help of Hamilton energy formula (57), the phase portraits with periodic potential wells behaviors for different value of $H_2$ are plotted in Fig. 18(b). It is found that there are the centers $S_2$ and the saddle $S_1$ for the unperturbed robot system (56).

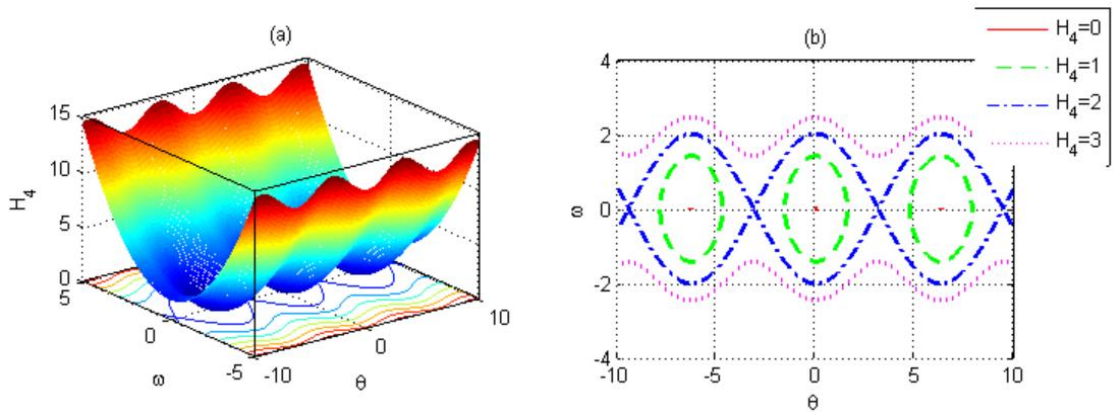

Fig. 18 Hamilton function $H_4$ surface and phase portraits of system (51). (a) Energy surface and (b) trajectories for the different values of Hamilton function $H_4$.

For the conservative form of the robot system (55) with parameter setting $\xi = 0$ and $M_0 = 0$, the analytical expression of the homo-clinic orbit are obtained as follows

$$\begin{cases} \theta_{\text{het1}}(T) = \pm 2\arctan(\sinh T) \\ \omega_{\text{het1}}(T) = \pm 2\operatorname{sech} T \end{cases} \quad (57)$$

where $T > 0$ is dimensionless time.

Making using of the heteroclinic orbit of Eq. (57), then the Melnikov function is computed by using the residues and obtained as following

$$M(T_0) = -\xi_0 \frac{4\pi^3}{3} \pm 2\pi \tanh\left(\frac{\pi\Omega_0}{2}\right)\sin(\Omega_0 T_0) \quad (58)$$

where $|\sin(\Omega_0 T_0)| \le 1$ and $\coth(*)$ is the hyperbolic cotangent function.

To find the criteria for the system (51), we start by setting the condition of simple zero $M(T_0) = 0$, one gets

$$M_0 > \frac{2\theta_3 \xi_0}{3\pi} \coth\left(\frac{\pi\Omega_0}{2}\right) = R_2(\xi_0, \Omega_0) \quad (59)$$

where $\theta_3$ refer to Eq. (16) and $\coth(*)$ is the hyperbolic cotangent function.

Fig. 19(a) shows the chaotic critical curved surface of system (56) in three dimensional parameters $\Omega_0 - \xi_0 - M_0$ space. The chaotic may happen when the parameters setting above this surface. In Fig. 19(b), the chaotic curves for different damping ratio $\xi_0 = 0.1, 0.2, 0.3$ and $0.4$. The chaotic behaviors will take place for non-dimensional force $F_0$ is the greater than the critical lines. Fig. 19(c) shows the chaotic threshold curves of system (51) for the different external periodic excitation frequency $\Omega_0 = 0.1, 0.2, 0.5$ and $1$. Fig. 19(d) presents the chaotic critical curves for the different external amplitude $M_0 = 0.2, 0.4, 0.6$ and $0.8$.

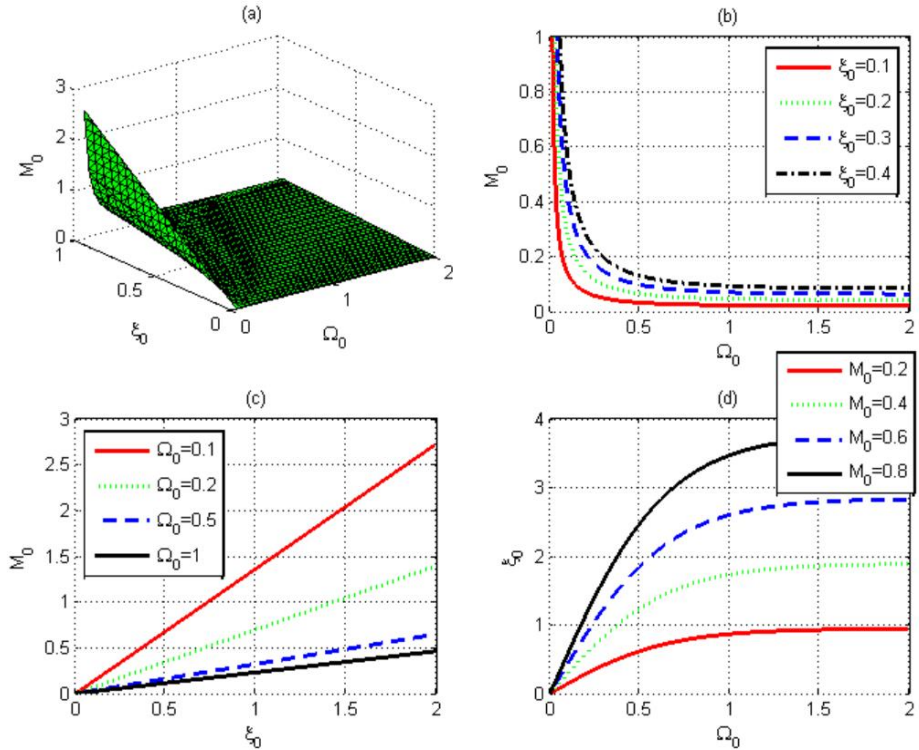

Fig. 19 Chaotic threshold of system (53). (a) Chaotic criteria surface and (b) curves for different value $\xi_0$. (c) Chaotic critical curves for different value $\Omega_0$. (c) Chaotic critical curves for different value $M_0$

### 5.3.3 Approximated approach of single well Duffing system with soft stiffness

Based on the topological equal technique, a tow dimensional first-order equations of unperturbed dipteran flight robot system obeys the differential equation

$$\begin{cases} \dot{\theta} = \omega \\ \kappa\dot{\omega} = -2\xi\omega - K\theta(\theta-\pi)(\theta+\pi) + M_0 \sin(\Omega_0 T) \end{cases} \quad (60)$$

where $\theta_1 = \pm\pi$ is the equilibrium angles as defined before.

Applied the integral methods, the Hamilton function for the autonomous robot system (61) is obtained as following

$$H_5 = \frac{1}{2}\omega^2 - \frac{1}{4}(\theta^2 - \pi^2)^2 \quad (61)$$

where $H_3$ is the total energy for the autonomous system (61).

As seen in Fig. 20(a), the Hamilton function surface with periodic potential wells is plotted. The undulating surface represents the Hamilton function of the total energy level, while the curves on the variation of $(\theta, \omega)$ the phase plane represent total energy levels. In Fig. 20(b), with the help of Hamilton energy formula (61), the phase portraits with periodic potential wells behaviors for different value of $H_2$ are plotted. It is found that there are the centers $S_2$ and the saddle $S_1$ for the unperturbed robot system (61).

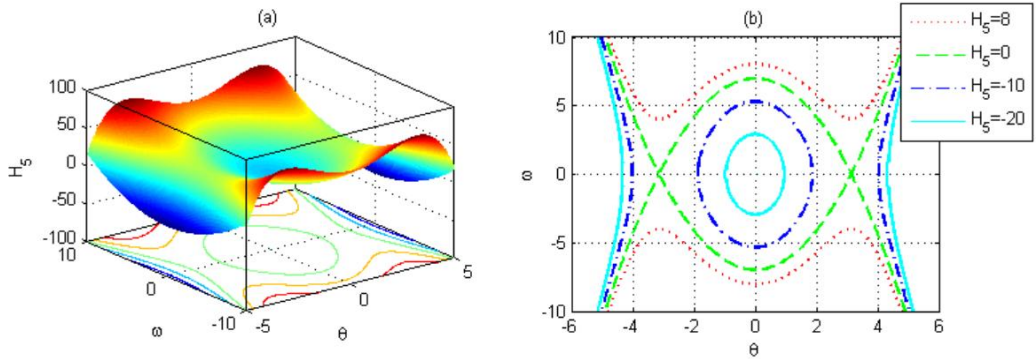

Fig. 20 Hamilton function $H_5$ surface and phase portraits of system (61). (a) Energy surface and (b) trajectories for the different values of Hamilton $H_5$.

The homo-clinic orbit are obtained in analytical form as follows

$$\begin{cases} \theta_{het2}(T) = \pm 2\tanh(T) \\ \omega_{het2}(T) = \pm 2\text{sech}^2 T \end{cases} \quad (62)$$

where $T > 0$.

The Melnikov function for system (60) is computed by using the heteroclinic orbit Eq. (62) and the residues theorem and written as following formula

$$M(T_0) = -\xi_0 \frac{4\theta_1^3}{3} \pm 2\pi\sinh\left(\frac{\pi\Omega_0}{2}\right)\sin(\Omega_0 T_0) \quad (63)$$

where $|\sin(\Omega_0 T_0)| \leq 1$ and csch($*$) is the hyperbolic cosecant function.

To find the criteria for the system (51), we start by setting the condition of simple zero $M(T_0) = 0$ with a simple zero and only if the following inequality holds

$$M_0 > \frac{2\theta_1 \xi_0}{3\pi}\text{csch}\left(\frac{\pi\Omega_0}{2}\right) = R_3(\xi_0, \Omega_0) \quad (64)$$

where csch($*$) is the hyperbolic cosecant function. Then the chaos will be generated when Melnikov boundary of Eq. (64) is beyond.

Fig. 21(a) shows the chaotic critical surface in three dimensional parameters $\Omega_0 - \xi_0 - M_0$ space. The chaotic may happen when the parameters setting above this surface. In Fig. 21(b), the chaotic curves for different damping ratio $\xi_0 = 0.1, 0.2, 0.3$ and $0.4$. The chaotic behaviors will take place for non-dimensional force $F_0$ the greater than the critical lines. Fig. 21(c) shows the chaotic threshold curves of system (51) for the different external periodic excitation frequency $\Omega_0 = 0.1, 0.2, 2$ and $5$. Fig. 21(d) presents the chaotic critical curves for different external amplitude $F_0 = 0.2, 0.4, 0.6$, and $0.8$.

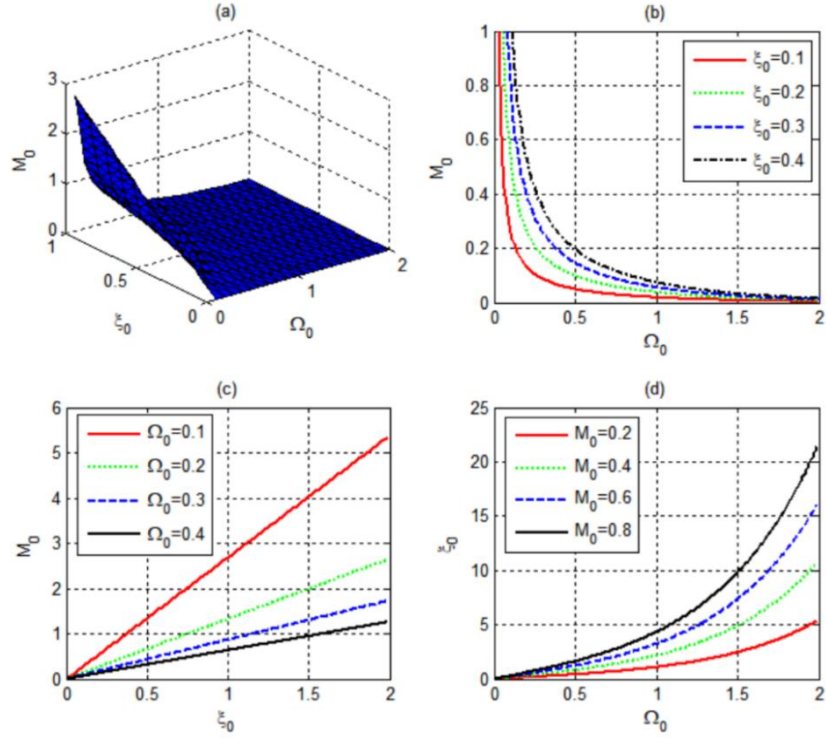

Fig. 21 Chaotic threshold of system (55). (a) chaotic criteria surface and (b) curves for different value $\xi_0$. (c) chaotic critical curves for different value $\Omega_0$. (d) Chaotic critical curves for different value $M_0$.

## 6. Experimental validation

Since the experimental verification is used to demonstrated the correction of the proposed robot design. The experimental platform on the vibration of the flapping robot has been established. The experimental study includes the following three aspects: (i) The static response of load of moment and angular discernment for the conservative system is carried out. (ii) The freely periodic vibration relationship between the amplitude and the frequency for the autonomous robot system is obtained. (iii) The forced oscillation of amplitude frequency relationship for perturbed robot system is given.

### 6.1 Experimental setup of the static moment analysis

As shown in Fig. 22(a), the prototype of flapping flight robot system with both nonlinear elastic restoring force and nonlinear damping force due to the geometrical construction are constructed. The scientific design size of flapping wing dynamic model imitating diperan flight with 470mm wing span, 350mm length and plastic material. This flapping robot model and flapping like a bird with a rubber band dynamic flapping wing. It is a assembled model, which has a more realistic shape, a simpler structure, and an enlarged wingspan and fuselage to comprehensively improve the time of stagnation and flight attitude. Moreover, the nose, tail and other parts have been improved, with a large wingspan and a more beautiful flight attitude. The moment of forces of the wings rotating the axis have be tested by handheld dynamometer. Additionally, the protractor is used to measure the angle displacement of the flapping wings.

As illustrated in Fig. 22(b), the prototype of flapping flight robot system with both nonlinear elastic restoring moment due to the geometrical construction are constructed. The relationship curves of the nonlinear moment of force $M_s$ versus the included angle $\theta$ are plotted for nine differential angle values of $\theta$, namely $\theta = -\pi, -0.75\pi, -0.5\pi, 0.25\pi, 0, 0.25\pi, 0.5\pi, 0.75\pi$ and $\pi$. It is found that the experimental data of the verification are in excellent agreement with the theoretical result Eq. (13).

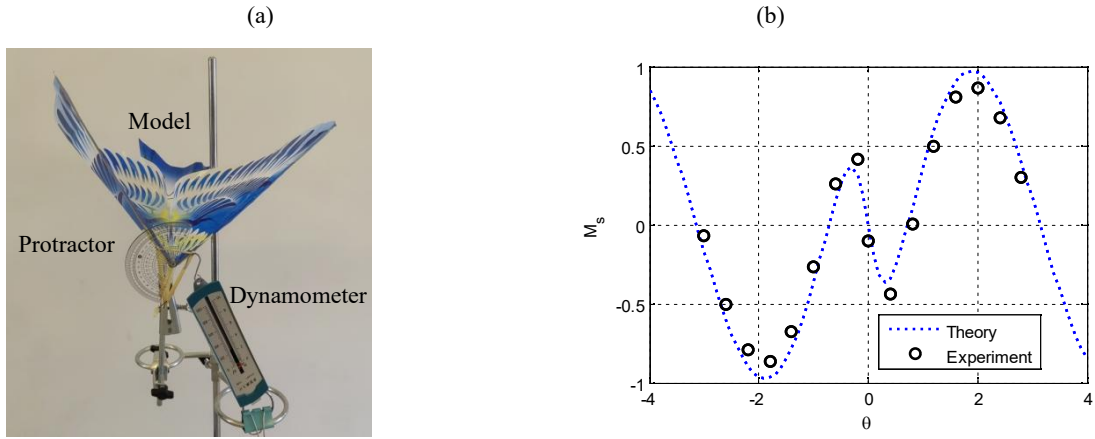

Fig. 22 Experimental setup of the static moment of force test. (a)Prototyping model and (b) the nonlinear moment of force with $\alpha = 1$ and $\beta = 1.5$.

### 6.2 Experimental test of freely periodic response

As shown in Fig. 23(a), the freely vibration prototype of the flapping dipteran robot system in the validation experiment was fabricated by mechanical fly joy with nonlinear elastic restoring moment. The detailed material parameter and geometrical properties are listed in Tab. 1. The sensor is used to test the flapping frequency. While, the camera is used to record the time history of vibration and measure the Instantaneous angle. In Fig. 23(b), the time history response is given by different initial value.

To study the effect of the oscillation characteristic variation on the relationship between the vibrating amplitude and the freely frequency, the freely vibration of the robot system is carried out experimentally. As depicted in Fig. 23(b), the amplitude frequency response are plotted for the different initial angle $\theta_0$ and the speed $\omega_0 = 0$. By comparison of the theoretical result, the numerical method and the experimental test, the proposed robot model is reliable and effective.

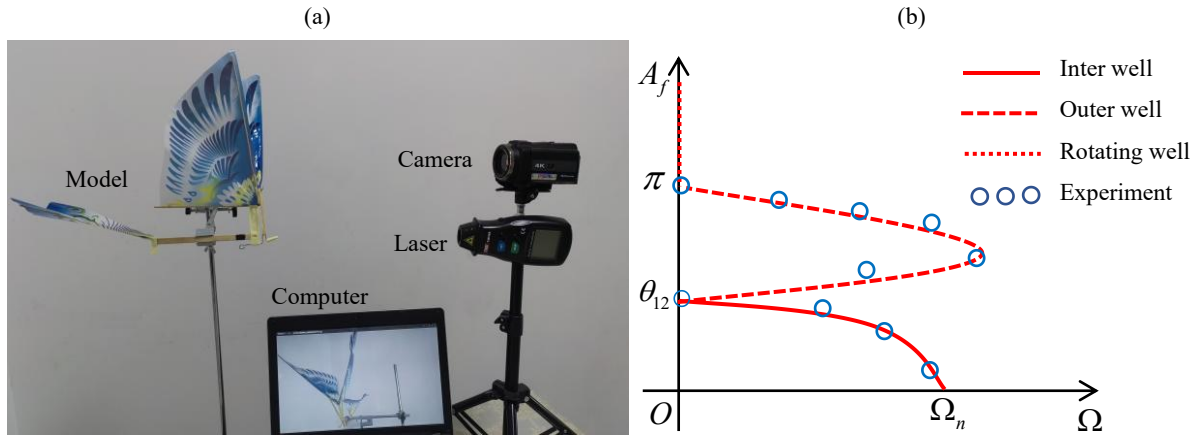

Fig. 23 The periodic response. (a) Experimental setup and (b) The double well with in region IV for $\alpha = 1$ and $\beta = 1.5$.

### 6.3 Experiment work of the forced periodic response

As shown in Fig. 24(a), the experimental vibration of the forced flapping dipteran robot system is fabricated with the external periodic moment. The flight mechanism of this robot is that two wings are flapped to produce the lift and thrust, which overcome the drag damping and gravitational force to provide the continuous flight. Moreover, the laser sensor is used to test the flapping frequency and the high speed camera is applied to record the time history of vibration and measure the instantaneous angle.

In Fig. 24(b), the amplitude frequency curves are plotted for the different excited frequency the speed $\Omega_0$. It is found that the nonlinearity bends to the left, that is similar to the soft Duffing system. The experiment of frequency of excitation is slowly varied up and down through the linear frequency is performed and the amplitude of the harmonic response is observed. Moreover, when the experiment can be started at pint A, and s is slowly decreased, a jump phenomenon take place from point C to point D. On the other hand, while the experiment started at point E with increased and the jump phenomenon occurs from point F to B. In meanwhile, it is shown that the experiment data, theory result and numerical response are consistent.

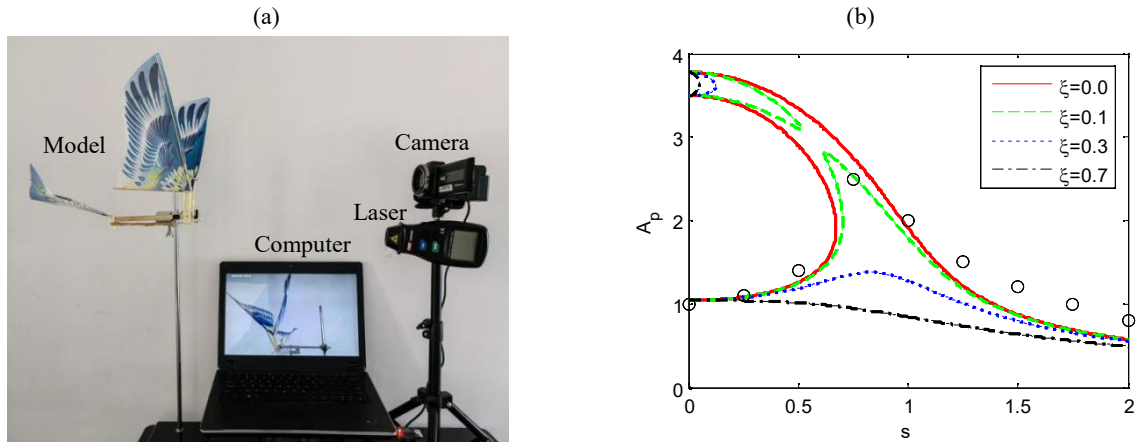

Fig. 24 Experimental setup. (a)Prototyping model and testing construction and (b) test results of amplitude response for different external excited frequency. ---- denoted the theoretical curves and ∘∘∘ represented the experimental results.

## 7. Conclusions

In this study, a novel construct design of the nonlinear fly robot system have been inspired by the flight mechanism of dipteran insects. The novel dipteran robot model with click mechanism is proposed based on the nonlinear geometrical mechanism of the rotating angle. The equation of motion for the flapping robot exhibits the fractional and radical nonlinearities. The nonlinear potentials, nonlinear elastic moments, and various phase portraits are investigated to display the smooth and nonsmooth characteristics depending the geometrical parameter $\alpha$ and $\beta$. The equilibrium bifurcations, nonlinear stiffness, as well as linearized equilibrium stability are studied, with detailed demonstrations of monostability and bistability. The amplitude frequency responses are obtained for the free vibration dynamic system with momonostability and bistability. Chaos thresholds are investigated using a topological equivalence method that overcomes the drawbacks of the Taylor expansion method. The prototype of flapping robot, the experimental apparatus and the test results are studied to verify the correctness and efficiency of the proposed the nonlinear flapping model.

On the basis of these studies, the following conclusions were drawn: (i)The theoretical, numerical and experimental results laying the foundation for the structural design of flapping wing flying robots [35]. (ii) The geometrical nonlinearity of an essential mechanism is critical in modelling of flight construction. This new typical design is used to improve the flight performance of dipteran robot [36]. (iii) The proposed methodology can be promoted to other engineering applications of vehicle shock absorption [37], energy harvesting [38].

**DRediT authorship contribution statement**

**Yanwei Han:** Conceptualization, Investigation, Writing-original draft. **Zijian Zhang:** Methodology, Funding acquisition, Writing review & editing.

**Declaration of Competing Interest**

The authors declare that they have no known competing financial interest or personal relationships that could have appeared to influence the work reported in this paper.

**Funding acknowledgement**

This work was supported by the State Key Laboratory of Robotics and System (HIT) (Grant no. SKLRS-2022-KF-19). Additionally, the authors would like to thank the anonymous reviewers and the editors for their relevant comments and useful suggestions.

**References**

[1] Ma KY, Chirarattananon P, Fuller SB, et al. Controlled flight of a biologically inspired, insect-scale robot. Science, 2013, 340: 603-607.

[2] https://www.britannica.com/summary/Leonardo-da-Vincis-Achievements#/media/1/336408/190393.

[3] Iwamoto H, Inoue K, Yagi N. Structure, function and evolution of insect flight muscle. Biophys J, 2010, 99: 21-28.

[4] Thomson AJ, Thompson WA. Dynamics of a bistable system: the click mechanism in dipteran flight. Acta Biotheor, 1977, 26 (1): 19-29.

[5] Brennan MJ, Elliott SJ, Bonello P, et al. The "click" mechanism in dipteran flight: if it exists, then what effect does it have? J Theor Bio, 2003, 224: 205-213.

[6] Lau GK, Chin YW, Goh JTW, et al. Dipteran-insect-inspired thoracic mechanism with nonlinear stiffness to save inertial power of flapping-wing flight. IEEE transactions on robotics, 2014, 30: 1187-1197.

[7] Gunther F, Shu Y, Iida F. Parallel elastic actuation for efficient large payload locomotion. IEEE International Conference on Robotics & Automation IEEE, 2015.

[8] Abas M, Rafie A, Yusoff HB, et al. Flapping wing micro-aerial-vehicle: Kinematics, membranes, and flapping mechanisms of ornithopter and insect flight. Chinese Journal of Aeronautics, 2016, 29: 1159-1177.

[9] Jankauski MA. Measuring the frequency response of the honeybee thorax. Bioinspiration biomimetics, 2020, 15: 046002.

[10] Lietz C, Schaber C F, Gorb S N, et al. The damping and structural properties of dragonfly and damselfly wings during dynamic movement. Communications Biology, 2022, 4: 737.

[11] Combes SA. Flexural stiffness in insect wings II. Spatial distribution and dynamic wing bending. Journal of Experimental Biology, 2003, 206: 2989-2997.

[12] Duncan W, Haldane, et al. Robotic vertical jumping agility via series-elastic power modulation. Science Robotics, 2016, 1: 2048.

[13] Hunt J, Giardina F, Rosendo A, Iida F. Improving efficiency for an open-loop-controlled locomotion with a pulsed actuation. IEEE/ASME Transactions on Mechatronics, 2016, 21: 1581-1591.

[14] Somps J, Lutges M. Dragonfly flight: novel uses of unsteady separated flows. Science, 1985, 228: 1326-1329.

[15] Warrick D, Tobalske B, Powers D. Aerodynamics of the hovering hummingbird. Nature, 2005, 435: 1094-1097.

[16] Bomphrey RJ, Nakata T, Phillips N, et al. Smart wing rotation and trailing-edge vortices enable high frequency mosquito flight, Nature, 2017, 544: 92-95.

[17] Muijres FT, Chang SW, Van WG, et al. Escaping blood-fed malaria mosquitoes minimize tactile detection without compromising on take-off speed. J Experimental Biology, 2017, 220: 3751-3762.

[18] Nakata T, Noda R, Liu H. Effect of twist, camber and spanwise bending on the aerodynamic performance of flapping wings. J Biomechanical Science and Engineering, 2018, 13: 17-00618.

[19] Johansson LC, Engel S, Kelber A, et al. Multiple leading edge vortices of unexpected strength in freely flying hawkmoth. Scientific Reports, 2013, 3: 3264.

[20] Lehmann FO. Neural control and precision of spike phasing in flight muscles. SciAccess Publishers, 2017, 2: 15-19.

[21] Lehmann FO, Wang H, Engels T. Vortex trapping recaptures energy in flying fruit flies. Sci Rep 2021, 11: 6992.

[22] Dickinson MH. Wing rotation and the aerodynamic basis of insect flight. Science, 1999, 284: 1954-1960.

[23] Chin YW, Kok JM, Zhu YQ, et al. Efficient flapping wing drone arrests high-speed flight using post-stall soaring. Sci Robot, 2020, 5: eaba2386.

[24] Heitler W J. The locust jump. Journal of Comparative Physiology A, 1974, 89: 93-104.

[25] Iwamoto H and Yagi N. The molecular trigger for high-speed wing beats in a bee. Science, 2013, 341: 1243-1246.

[26] Ilton M. Bhamla S, Ma XT, et al., The principles of cascading power limits in small, fast biological and engineered systems. Science, 2018, 360: eaao1082.

[27] Steinhardt E, Hyun NP, Koh JS, et al. A physical model of mantis shrimp for exploring the dynamics of ultrafast systems. Proc Natl Acad Sci, 2021, 118: e2026833118.

[28] Harne RL, Wang KW. Dipteran wing motor-inspired flapping flight versatility and effectiveness enhancement. J R Soc Interface. 2015, 12: 20141367.

[29] Majumdar D, Ravib S, Sarkar S. Passive dynamics regulates aperiodic transitions in flapping wing systems. PNAS Nexus, 2023, 2: 1-10.

[30] Wiggins S. Introduction to Applied Nonlinear Dynamical System and Chaos. New York: Springer-Verlag, 1990.

[31] Globitsky M, Sterwart I, Schaeffer DG. Singularities and groups in bifurcation theory. New York: Springer-Verlag, 1988.

[32] Nayfeh AH, Mook AD. Nonlinear Oscillations. New York: John Wiley & Sons, 1995.

[33] Guckenheimer J, Holmes P. Nonlinear Oscillations, Dynamical Systems, and Bifurcations of Vector Fields, New York: Springer-Verlag, 1983.

[34] Rao SS. Mechanical Vibration. New York: Prentice Hall, 2004.

[35] Lehmann FO. Neural control and precision of spike phasing in flight muscles. Journal of Neurology & Neuromedicine, 2017, 2:15-19.

[36] Chi YD, Li YB, Zhao Y, et al. Bistable and multistable actuators for soft robots: structures, materials, and functionalities. Adv Mater, 2022, 34: 2110384.

[37] Yu M, Arana C, Evangelou SA, et al. Quarter-car experimental study for series active variable geometry suspension. IEEE Transactions on Control Systems Technology, 2017, 27: 743-759.

[38] Xu GQ, Li CY, Chen CJ. Dynamics of triboelectric nanogenerators: A review. Int J Mech Sys Dyn, 2022, 2: 311-324.